\documentclass[10pt,twocolumn,letterpaper]{article}

\usepackage{cvpr}              

\usepackage[accsupp]{axessibility} 
\usepackage{graphicx}
\usepackage{amsmath}
\usepackage{amssymb}
\usepackage{epigraph} 
\usepackage{booktabs}
\usepackage{float}
\usepackage{dblfloatfix}
\usepackage{multirow}
\usepackage{xcolor}
\usepackage[breaklinks,colorlinks]{hyperref}

\usepackage[capitalize]{cleveref}
\crefname{section}{Sec.}{Secs.}
\Crefname{section}{Section}{Sections}
\Crefname{table}{Table}{Tables}
\crefname{table}{Tab.}{Tabs.}

\def\paperID{6040} 
\def\confName{CVPR}
\def\confYear{2024}

\begin{document}

\title{Weak-to-Strong 3D Object Detection with X-Ray Distillation}


\author{Alexander Gambashidze\thanks{Corresponding author, email: alexandergambashidze@gmail.com}  \textsuperscript{ 1,2} \and Aleksandr Dadukin\textsuperscript{2} \and Maxim Golyadkin\textsuperscript{1, 2} \and Maria Razzhivina\textsuperscript{2} \and Ilya Makarov\textsuperscript{1,3}}

\maketitle

\addtocounter{footnote}{1}
\footnotetext[1]{Artificial Intelligence Research Institute}
\addtocounter{footnote}{1}
\footnotetext[2]{HSE University}
\addtocounter{footnote}{1}
\footnotetext[3]{ISP RAS}

\begin{abstract}

This paper addresses the critical challenges of sparsity and occlusion in LiDAR-based 3D object detection. Current methods often rely on supplementary modules or specific architectural designs, potentially limiting their applicability to new and evolving architectures. To our knowledge, we are the first to propose a versatile technique that seamlessly integrates into any existing framework for 3D Object Detection, marking the first instance of Weak-to-Strong generalization in 3D computer vision. We introduce a novel framework, X-Ray Distillation with Object-Complete Frames, suitable for both supervised and semi-supervised settings, that leverages the temporal aspect of point cloud sequences. This method extracts crucial information from both previous and subsequent LiDAR frames, creating Object-Complete frames that represent objects from multiple viewpoints, thus addressing occlusion and sparsity. Given the limitation of not being able to generate Object-Complete frames during online inference, we utilize Knowledge Distillation within a Teacher-Student framework. This technique encourages the strong Student model to emulate the behavior of the weaker Teacher, which processes simple and informative Object-Complete frames, effectively offering a comprehensive view of objects as if seen through X-ray vision. Our proposed methods surpass state-of-the-art in semi-supervised learning by 1-1.5 mAP and enhance the performance of five established supervised models by 1-2 mAP on standard autonomous driving datasets, even with default hyperparameters. Code for Object-Complete frames is available here: https://github.com/sakharok13/X-Ray-Teacher-Patching-Tools.
\vspace{\parskip}
\end{abstract}

\begin{figure}[H]
    \centering
    \includegraphics[width=\linewidth]{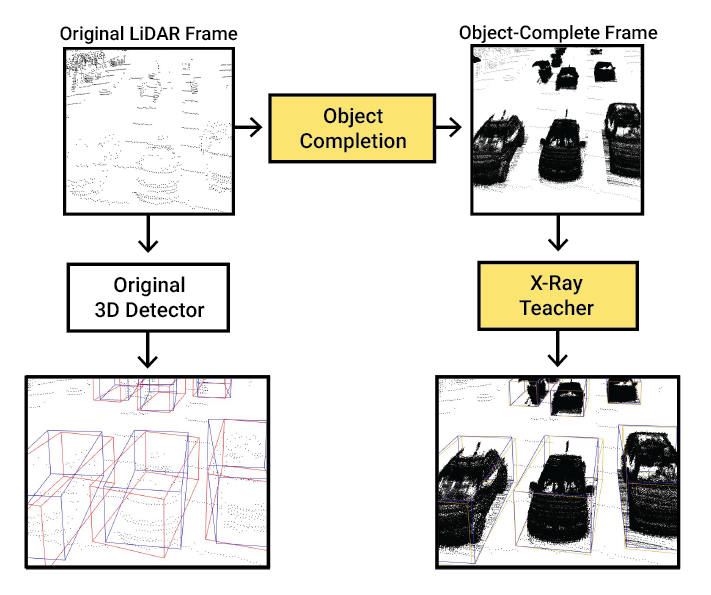}
    \caption{3D object detection directly from sparse LiDAR data (top left) provides noisy predictions (bottom left). Adding object completion stage (top right) helps to train 3D object detection X- Ray Teacher, which is robust and can be distilled to baseline model. Red, Yellow and Blue colors of bounding boxes are related to classical LiDAR-based object detection, Our model on Object-Complete frames predictions and Ground Truth labels, respectively.  
    }
    \label{fig:enter-label}
\end{figure}

\epigraph{``You're just not thinking fourth dimensionally... \textit{the bridge will exist.}''}{\textit{ --- Dr. Emmett Brown, "Back to the Future III"}}

\section{Introduction}
\label{sec:intro}


3D object detection is a fundamental task in the field of computer vision and autonomous systems, playing a key role in the advancement of self-driving technology \cite{selfdriving1} and contributing significantly to the robotics industry \cite{robotics1, robotics2}. Currently, LiDAR-based 3D object detection \cite{lidar1, lidar2, dsvt} demonstrates superior performance compared to camera-based \cite{camera1, camera2, camera3, camera4} and radar-based \cite{radar3, radar4, radar2, radar1} approaches. Furthermore, LiDAR point clouds are the key ingredient of multimodal fusion-based approaches \cite{fusion1, fusion2, fusion3}, so LiDAR-based 3D detection continues to be a strong focus of the research community.

The point cloud-based 3D object detection challenges the following issues: sparsity, occlusions, and the complexity of 3D data annotation. \textbf{Sparsity}: large point clouds are sparse due to the LiDAR sensing process's inherent characteristics that leads to an imprecise representation of the captured scene. Additionally, there is an imbalance in point density. In particular, the point cloud is sparser in the far range and contains less spatial information affecting feature representation and box prediction. \textbf{Occlusions}: another problem arises from the frequent occurrence of partial occlusion in LiDAR frames. This primarily happens due to the fact that the frames are acquired from a single fixed point of view, making them essentially 2.5D. To detect and accurately locate a highly occluded object, a detector must recognize the hidden shapes of the object even when a significant portion of its parts is missing. Since the absence of certain shapes inevitably affects object perception, this becomes a critical detection challenge. Our previous approaches focused on depth completion \cite{makarov2017semi}, depth inpainting \cite{semenkov2024inpainting} or self-supervised depth pretraining \cite{karpov2022exploring,indyk2023monovan} could not address occlusion problems. \textbf{Data annotation}: finally, data annotation is a formidable challenge in 3D object detection due to the complexity of annotating objects in three-dimensional space. For example, a skilled annotator can spend weeks annotating just one hour of LiDAR data \cite{annots1, once}. This problem is partially addressed by semi-supervised learning approaches in a teacher-student framework \cite{MeanTeachers, 3diou, ProfTeachers, paper_ses}. However, the quality of pseudo labels and the performance of these methods are limited due to the aforementioned sparsity and occlusion problems. Therefore, overcoming these inherent challenges is essential to improve the accuracy and reliability of 3D detection systems.

The challenges posed by sparsity and occlusion have led to the formulation of several methodologies. Some of them \cite{second} are aimed at improving the computational efficiency of processing sparse data without improving the quality of the predictions. Others, including our work, focus on improving detection performance on sparse occluded data \cite{dops, sienet, btc, ddit}. Li et al. \cite{ddit} enforce shape constraints to improve object localization with explicit shape priors obtained from a database of CAD models. Najibi et al. \cite{dops} and Li et al. \cite{sienet} do the same by implicitly introducing priors with novel modules pretrained for point cloud completion and SDF approximation. Xu et al. \cite{btc} propose a Shape Occupancy Probability estimation module to refine predicted bounding boxes for occluded objects. However, all of these mentioned approaches \textit{utilize new modules and adapt the model architecture to take advantage of them}. As a result, these methods have limited applicability to new architectures that will emerge as the field evolves. 

In this work,  \textit{we address all three major challenges in 3D object detection} with our novel framework called \textit{X-Ray Distillation} with Object-Complete Frames that is easily plugged in existing approaches and new architectures. It is designed for universal application to any LiDAR-based detector, improving performance on sparse and occluded objects. Our approach exploits the properties of existing large-scale autonomous driving datasets, which consist of sequences of LiDAR frames. Such property makes it possible to reconstruct complete shapes for occluded objects using other occurrences of these objects in the sequence, ensuring that all objects are equipped with points from all available viewpoints in a scene. We then use this completed data in the Teacher-Student framework for both semi-supervised learning and knowledge distillation in a supervised setting. We train our Teacher on extremely informative Object-Complete frames thus making it a weaker model \cite{burns2023weak}. Then, we use it to extract features from such simple Object-Complete frames and distills this knowledge to a stronger Student, which operates with the original data, to guide him on how to extract rich features from occluded objects. To generate Object-Complete frames, we leverage ground truth object tracking labels. Since there are no labels for the majority of data in the semi-supervised setting, we propose an Objects Temporal Fusion block to detect, track, and use point cloud registration techniques to construct Object-Complete frames.


We validate the proposed X-Ray Teacher framework on nuScenes \cite{NuScenes} and Waymo Open Dataset \cite{waymo} 3D object detection benchmarks for SECOND \cite{second}, CenterPoint \cite{centerpoint}, and DSVT \cite{dsvt} models in a supervised learning paradigm. Experiments show a steady improvement by 1-2 mAP with minimal or no impact on time and computational resources during the inference stage. For semi-supervised performance evaluation, we use ONCE \cite{once} benchmark.

Applying these novel ideas for 3D object detection, we provide the following contributions:
\begin{enumerate}
     \item We propose the X-Ray Teacher framework for semi-supervised learning, which achieves state-of-the-art performance on the ONCE benchmark.
    \item We show that our approach improves the quality of four supervised learning models, including the current state-of-the-art model, and demonstrates the potential to improve the performance of any supervised model trained on sequential data.
    \item We suggest the Objects Temporal Fusion block to generate Object-Complete frames for data that lacks ground truth tracking labels.

\end{enumerate}

\section{Related work}

\label{related_works}

\subsection{Object Detection for point clouds}
Modern 3D detectors predominantly use point clouds, voxel representations, or a combination of both. The pioneering PointNet model family \cite{pointnet, pointnetpp} provided a significant step forward in 3D recognition. However, direct processing of point clouds poses several challenges: it requires point sampling, grouping, and computation of point-wise features, which can be computationally intensive. To integrate insights from 2D computer vision, a transition to a pixel equivalent, namely voxels, is required. Methods such as \cite{pixor, centerpoint, centerformer, btc} first convert point clouds to voxels, followed by the use of 3D sparse convolutions \cite{spconv}. Recent advances also include the integration of modified self-attention layers, achieving state-of-the-art results \cite{dsvt}.

\subsection{Semi-supervised 3D detection}
Although Semi-supervised 3D object detection is not as extensively researched as 2D detection, it has seen some significant contributions. Among these, SESS \cite{paper_ses} stands out for using the general Mean Teacher approach with data augmentations and consistency loss. Similarly, 3DIoUMatch \cite{3diou} is notable for its unique localization strategies and 3D IoU-guided techniques for box filtering. Proficient Teachers \cite{ProfTeachers} is notable for its box voting and contrastive losses. We do not propose yet another standalone method for Semi-supervised 3D Object Detection; rather, our emphasis is on creating a plug-in technique designed to augment and enhance the performance of existing methods.

\subsection{Knowledge Distillation}
The concept of Knowledge Distillation (KD) was first introduced by Hinton et al. \cite{Hinton}. KD describes a learning approach where a larger teacher network guides the training of a smaller student network for various tasks \cite{KD}. Broadly, KD methodologies are categorized into two types: logits/regression distillation and feature map distillation. Our focus is on a hybrid approach, combining these methods: matching feature maps, regression, and classification heads with pseudo labels generated by the teacher model.

In 3D object detection, Knowledge Distillation is mainly utilized to minimize parameters and FLOPS while aiming to preserve box prediction quality \cite{distill1, distillation}. Yet, methods that aim to outperform state-of-the-art models are uncommon. In contrast to this trend, our X-Ray Teacher model challenges the norm by providing the quality improvement.

\begin{figure*}[btp]
    \centering
    \includegraphics[width=.9\linewidth]{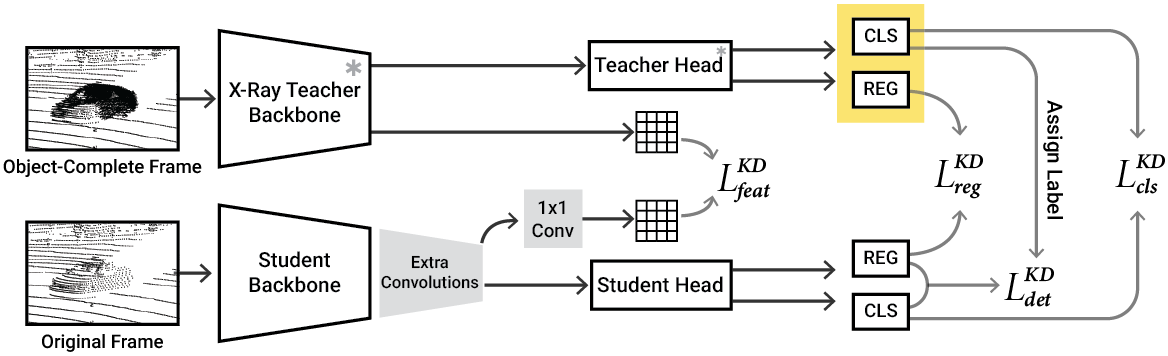}
    \caption{Overall X-Ray Knowledge Distillation for Supervised Learning.
    X-Ray Teacher is frozen and pretrained on Object-Complete frames which are taken as input. The Student is guided to mimic the Teacher's behaviour through Knowledge Distillation losses: $L_{feat}$ for intermediate embeddings matching, $L_{reg}$ for bounding box regression, $L_{det}$ for basic detection, and $L_{cls}$ for classification.}
    \label{fig:dist_pipeline}
\end{figure*}

\section{Methodology}
\label{methodology}


\subsection{Overview of X-Ray Teacher}
\label{overview}

We introduce a novel training framework to address the challenges of sparsity and occlusion in 3D Object Detection based on LiDAR data. This framework is not limited to any specific object detection model and has the potential for applications across various deep learning architectures. Our method is designed to process the LiDAR data structured as a sequence of frames.

The two core elements of our approach are Object Complete Frames Generation and Teacher-Student Knowledge Distillation. Our approach for 3D object detection can be applied in both, supervised and semi-supervised settings, with minor differences in the elements implementation.

\textbf{Object Complete Frames Generation}. In this step, we reconstruct the complete shapes of objects presented in the scene by utilizing information from the other frames within the same sequence. Given that autonomous driving datasets are composed of sequential data, we can efficiently leverage their temporal nature: we add points from both the future and the past when objects are observed from different viewpoints. It allows us to reconstruct the complete shapes of objects without shape databases or reconstruction modules.

In order to verify the validity of our approach, we trained a CenterPoint \cite{centerpoint} model on both, the original and object-complete NuScenes \cite{NuScenes} datasets, and then evaluated their performance on the respective validation sets. 
The models trained on the original and the object-complete frames achieved mAP scores of 59.2\% and 79.5\%, respectively. This difference of 20 mAP suggests that 1) it would be beneficial to transform unlabeled original point clouds into Object-Complete ones and to annotate them with a corresponding X-Ray Teacher pretrained on such informative frames 2) the features extracted by a weaker X-Ray Teacher might be distilled to a stronger student to share the knowledge of complete shapes.

    
    

\textbf{Teacher-Student Knowledge Distillation}. The necessity for this step arises because we cannot generate Object Complete Frames during the online inference stage, as it is not possible to access future data. Therefore, we need to encourage the model to behave as if it were observing shape-complete objects, even when dealing with occluded ones. A well-known method for enabling a deep learning model to imitate another model's behavior involves using Knowledge Distillation within a Teacher-Student framework. However, for conventional Knowledge Distillation, both the Teacher and Student models typically process data of the same complexity (the only difference may lie in the complexity of augmentations \cite{fixmatch}). In contrast to the standard knowledge distillation, we enrich the data for training Teacher, which significantly improves its performance in 3D object detection. Then, we teach the Student to extract important information from less detailed data by distilling knowledge from the Teacher model.

Instead of simplifying the Student model, as it is usually done in standard Knowledge Distillation, we take an opposite approach and design the Student to be more complex than the Teacher. It helps to extract high-quality information from more intricate and ambiguous data and demand a Student model to be more robust and to have a more complex receptive field capacity.

In what follows, we provide detailed descriptions of how we implement Object Complete Frames Generation and Teacher-Student Knowledge Distillation steps in both supervised and semi-supervised settings.

\subsection{Supervised X-Ray Teacher}
\label{supervised}

In the supervised setting of 3D Object Detection, models are trained and evaluated using datasets that provide labeled data, including precise bounding boxes and instance IDs. Object-Complete Frame Generation for labeled data involves aggregating objects based on their instance IDs and merging different views into a unified point cloud (see Section \ref{object_complete} for details).

For the distillation process (see Figure \ref{fig:dist_pipeline}), we train Teacher model on Object-Complete frames and then freeze it. Then, we train baseline model (playing Student role) to directly minimize Knowledge Distillation losses inspired by \cite{distillation}. The distillation is done by matching Teacher and Student backbone encoders' embeddings, output labels for bounding box regression, classes distribution for classification task (objects like pedestrians, cars, cyclists, etc.), and intermediate features obtained from the outputs of regression and classification heads before  postprocessing (assigning labels).
Specifically, we define the following losses: 
\begin{align}
    \mathcal{L}^{KD}_{\text{heads}} &= \alpha_1 \mathcal{L}^{KD}_{\text{reg}} + \alpha_2 \mathcal{L}^{KD}_{\text{cls}} \\
    &=\alpha_1  D_{KL}(S_{\text{cls}} \| T_{\text{cls}}) +\alpha_2 \text{MSE}(S_{\text{reg}}, T_{\text{reg}}) \\
    \mathcal{L}^{KD}_{\text{feat}} &= \text{MSE}(T_{\text{back}},  \phi(\omega(S_{\text{back}}))) \\
    \mathcal{L}^{KD}_{\text{det}} &= \mathcal{L}_{detection}(S_{preds}, \tilde{T}_{boxes})
\end{align}

 $T$ and $S$ are the outputs of our Teacher and Student models, respectively. The Student takes the original frame $F$ as input, while the Teacher receives the Object-Complete Frame $\tilde{F}$, so $F \subset \tilde{F}$. $S_{back}$ and $T_{back}$ are referred to the output of the backbone module and $S_{reg}, T_{reg}$, $S_{cls}, T_{cls}$ are the outputs of regression and classifications heads. $\tilde{T}_{boxes}$ are the X-Ray Teacher's predicted boxes after postprocessing. $S_{preds}$ is an overall Student's output; $\alpha_1$, $\alpha_2$ are non-negative hyper-parameters. $L_{detection}$ is a basic detection loss that is used for training 3D object detection models. $\phi$ is a 1x1 Convolution to better match the Teacher's feature maps. $\omega$ refers to some extra convolutions to make the Student more flexible. We discovered that X-Ray Distillation does not need extra convolutions in the case of the NuScenes dataset, so their usage for encoder feature adjustment is also a hyperparameter of the model. By MSE, we mean Mean Squared Error.

Finally, the training objective can be written as follows: 

\begin{align}
    \mathcal{L} = \lambda_1 \mathcal{L}^{KD}_{\text{heads}} + \lambda_2 \mathcal{L}^{KD}_{\text{feat}} + \lambda_3 \mathcal{L}^{KD}_{\text{det}}
\end{align}

where $\lambda_1$, $\lambda_2$, $\lambda_3$ are non-negative hyper-parameters balancing the contribution of each term.
\subsection{Semi-supervised X-Ray Teacher}
\label{semisupervised}

Semi-supervised learning is characterized by the availability of a small amount of labeled data and a larger pool of unlabeled data, making the use of the Object Complete Frame generation approach proposed for supervised settings infeasible. 

\begin{figure}[hbt!]
    \centering
    \includegraphics[width=\linewidth]{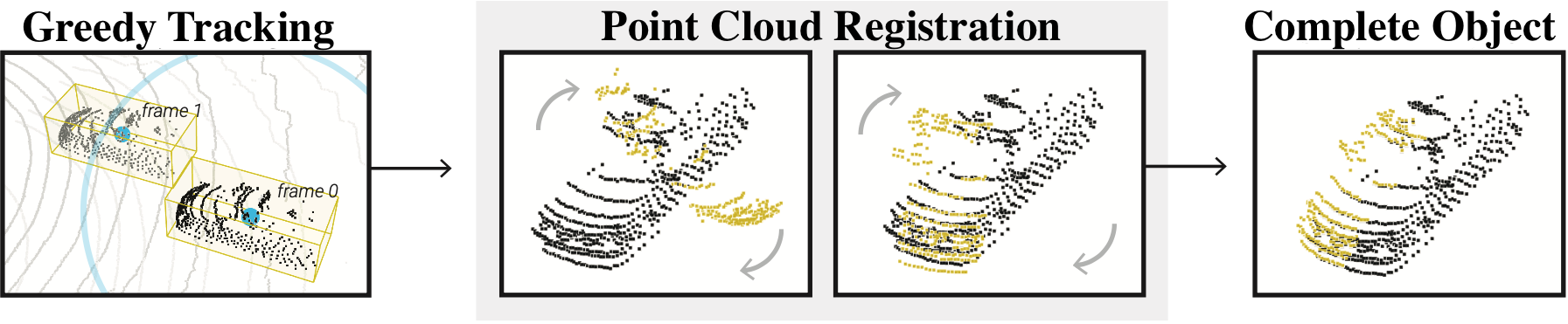}
    \caption{Object-Complete Frame generation process for semi-supervised setting. It consists of tracking and Point Cloud Registration. We track objects across all frames in the whole sequence, then we use Point Cloud Registration to merge points that represent the same object from different views, and finally we replace the original object with the new, complete one.
    }
    \label{otf}
\end{figure}

\begin{figure*}[hbt!]
    \centering
    \includegraphics[width=0.85\linewidth]{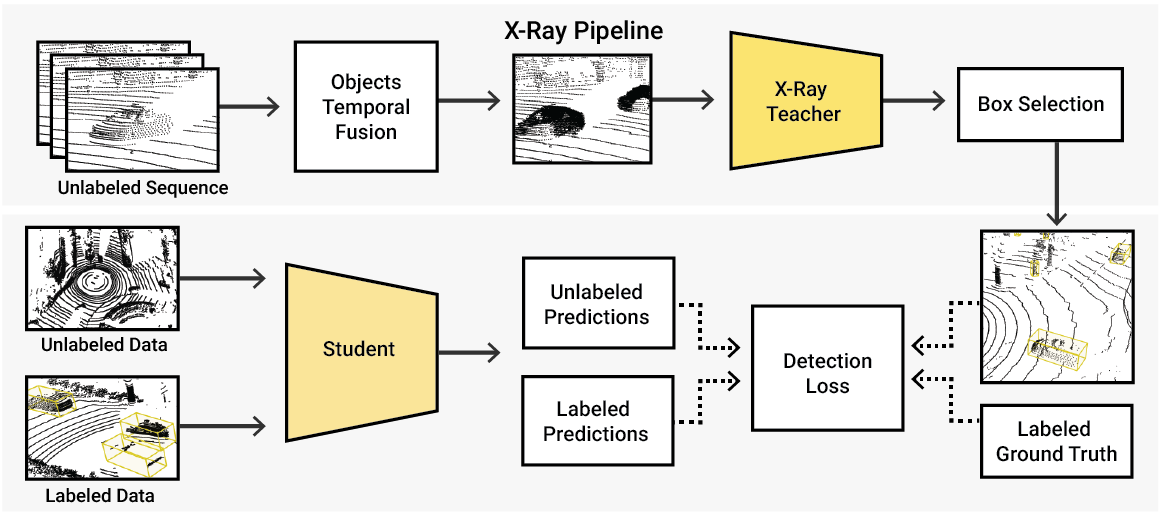}
    \caption{Semi-supervised X-Ray Teacher pipeline for 3D object detection task. Unlabeled sequences are processed by the Objects Temporal Fusion block to create more complete object representations by aggregating information over time. The Student model learns from both pseudo-labeled predictions and actual labeled data with ground truth annotations.}
    \label{fig:pipeline}
\end{figure*}

In order to overcome this limitation, we introduce the Objects Temporal Fusion block (as shown in Figure \ref{otf}), which is designed to enable Object Complete Frame generation in situations where ground truth labeling is missing. This block leverages a model pretrained on labeled data to detect and track objects across unlabeled sequences. Subsequently, it employs Point Cloud Registration (PCR) to merge objects points from different views for all detected objects.

The provided steps outline a precise and detailed procedure for Objects Temporal Fusion:
\begin{enumerate}
    \item to label all LiDAR frames using a pretrained model; 
    \item to greedily track objects across all frames in each sequence using predicted bounding boxes; assign unique IDs to every object instance to facilitate the identification of recurring objects within the scene and to organize them in a sequential manner;
    \item for each object within each sequence, to merge different views of the same object by applying a deep learning model for Point Cloud Registration; this process generates a complete point cloud for each object, which is then used to replace the occluded one in each frame;
    \item to fine-tune the base model on Object-Complete labeled frames; this refined model forms our X-Ray Teacher.

\end{enumerate}

More details for greedy tracking and Point Cloud Registration models can be found in Section \ref{object_complete}.

The Knowledge Distillation step presented in Figure \ref{fig:pipeline} can be integrated with any semi-supervised 3D Object Detection approach that uses pseudo labeling as a form of self-distillation. All methods in the domain usually follow this paradigm, which underlines the high universality of our approach. From this perspective, the use of Object-Complete frames for pseudo label prediction refines the 3D bounding boxes and improves the quality of self-distillation. The sole limitation is that existing semi-supervised methods \cite{MeanTeachers, ProfTeachers} typically update the Teacher model's weights using the exponential moving average of the Student's weights. We refrain from this practice because our Teacher is finetuned on Object-Complete frames and thus predicts higher-quality labels (see proofs in Section \ref{ablation_study}).

\section{Experiments}
\label{experiments}

This section covers the datasets, implementation details, experiment results, and a detailed analysis of the different components that affect performance. It starts with Section \ref{data}, which defines the datasets and metrics used to prove the validity of our ideas. Section \ref{details} delves into the network architectures and training parameters, while Section \ref{results} analyzes the obtained metric values. Finally, Section \ref{ablation_study} examines various parts of our approach and their influence on the overall performance.

\subsection{Data}
\label{data}

To evaluate our models, we use three large-scale autonomous driving datasets: NuScenes and Waymo Open Dataset for the supervised setting and ONCE for the semi-supervised setting.

\textbf{NuScenes} \cite{NuScenes} is a popular outdoor dataset with diverse annotations for different tasks. It has 40,157 annotated samples. For 3D object detection, we provide NuScenes Detection Score (NDS) and mean Average Precision (mAP).

\textbf{Waymo Open} \cite{waymo} is also one of the most popular outdoor 3D perception datasets. It contains 1150 point cloud sequences and has more than 200K total frames. All results are evaluated with 3D mean Average Precision (mAP) and its weighted variant (mAPH)

\textbf{ONCE} \cite{once} is a large-scale dataset with 1 million pout cloud samples from LiDAR and only 15K annotated frames that are divided into train, val, and test with 5k, 3k, and 8k samples, respectively. This dataset is designed exactly for semi-supervised learning tasks and simulates real life: annotations are expensive and time-consuming. All not labeled frames are divided into Small (70 sequences), Medium (321 sequences), and Large (560 sequences) parts. We follow the ONCE Benchmark and use mAP over all classes with the 3D IoU thresholds 0.7, 0.3, and 0.5 for classes "Vehicle", "Pedestrian", and "Cyclist", respectively.

\subsection{Implementation Details}
\label{details}

\subsubsection{Network architectures} 



For supervised setting, we use SECOND \cite{cbgs}, CenterPointVoxel \cite{centerpoint}, CBGS \cite{cbgs}, DSVT \cite{dsvt} within our framework. The implementations of these models are based on the OpenPCDet \cite{openpcdet2020} library, and we adhere to the default configurations suggested by this library for both training and inference. In addition, we present scaled versions of these models, which are essentially the original networks augmented with five additional convolutional layers stacked on top of the BEVEncoder with the following parameters 1x Conv(512, 128), 3x Conv(128, 128), and 1x Conv(128, 512) with Batch Normalizations and ReLU activations after each convolution. With the help of light Grid Search, we choose the following distillation hyperparameters: $\alpha_1 = 2$, $\alpha_2 = 1$, $\lambda_1 = 0.7$, $\lambda_2 = 0.3$, $\lambda_3 = 1$.

For semi-supervised setting, we follow the previous works \cite{3diou, ProfTeachers} and use SECOND and CenterPointVoxel models from OpenPCDet for validation of comparison. For training and inference, we use recently proposed refined configurations from work \cite{once_fix}. X-Ray Teacher model is fine-tuned on object-complete frames with the same hyperparameters for 10 epochs.

For computations, we use 4x A100 40GB GPU and AMD EPYC 7702 CPU.

\begin{table}[htb]
    \centering
    \caption{Evaluation of the impact of teacher fine-tuning on Object-Complete frames in semi-supervised setting on ONCE dataset. The results indicate that teacher fine-tuning is essential.}
    \begin{tabular}{l|l}
    \hline
    \textbf{Method} & \textbf{mAP} \\
    \hline
    X-Ray MT SECOND & 60.12\\
    X-Ray PT SECOND & 61.48\\
    X-Ray MT SECOND (fine-tuned) & 65.26 \\
    X-Ray PT SECOND (fine-tuned) & \textbf{68.43}\\
    \hline
    X-Ray MT CenterPoint & 59.35\\
    X-Ray PT CenterPoint & 62.17\\
    X-Ray MT CenterPoint (fine-tuned) & 67.60 \\
    X-Ray PT CenterPoint (fine-tuned) & \textbf{70.55}\\
    \hline
    \end{tabular}

    \label{tab:finetuning_comparison}
\end{table}

\subsubsection{Object-Complete Frame Generation}

\label{object_complete}

The key concept of this paper - Object-Complete Frame Generation - combines diverse ideas to achieve optimal Object Completion, encompassing facets such as detection, tracking, and the registration of point clouds.

The detection phase is construed as an elective stage, exclusively implemented on unlabeled data. Within this stage discerned instances are encapsulated into generated 3D bounding boxes. E.g., a substantial proportion of instances within the ONCE dataset lacks accompanying labels.

Subsequently, the greedy tracking procedure traverses the entire set of frames, associating the appearance of instances at the i-th frame with potential candidate appearances in the following frame. The list of candidate instances is built by including all instances from the succeeding frame that fall within a prescribed radius, defined as twice the maximum dimension across their respective bounding boxes. The nearest instance is selected from the list, while the remaining instances are discarded. The series of matches made using this algorithm are combined into a single sequence, called track. When there are no more potential matches in the next frame, the track is considered as terminated. For the given instance in some specific frame there is the only corresponding track across the entire scene.

Those prepared instances are then used for Object Completion, a process executed through several sequential steps:
\begin{enumerate}
    \item Point clouds associated with instances are extracted from their respective frames, they are translated back to the zero-point of the global basis. The rotation of the instances are also reset to identity.
    \item The point clouds corresponding to instances within a common track undergo a merging process, constituting the Point Cloud Registration phase, wherein diverse set of merging approaches is used to glue them into a larger, densely populated point cloud.
    \item Later on the corresponding point clouds are replaced with their respective densely populated point cloud from the previous step, restoring their original translation and rotation in a specific frame.
\end{enumerate}

\begin{table}[htb]
    \centering
    \caption{Comparison of three distillation strategies: using only classification and regression heads matching, BEV features and heads matching, and the final pipeline that is the SECOND-Scaled model; experiment performed using Waymo Validation set.}\label{tab:distillation_comparison}
    \resizebox{\columnwidth}{!}{
    \begin{tabular}{l|l|l}
    \hline
    \textbf{Technique} & \textbf{mAP/mAPH L1 } & \textbf{mAP/mAPH L2}\\
    \hline
    Heads match & 67.5/63.4 & 61.1/57.2 \\
    BEV \& Heads & 67.8/63.5 & 61.6/57.4  \\
    Full Pipeline (ours) & \textbf{68.3}/\textbf{64.3} & \textbf{61.9}/\textbf{58.0}  \\
    \hline
    \end{tabular}}   
\end{table}

Various merging strategies were empirically tested in generating Object-Complete point clouds:

\begin{enumerate}
    \item Geometry: assumes the objects (with boxes) are best aligned by the detected bounding box, performs both inverse translation and rotation to clear their geometric transformations and then merges intact point clouds.
    \item GeDi: uses GeDi Point Cloud Registration method \cite{gedi}.
    \item Greedy Grid: uses Greedy Grid implementation \cite{Bojanić-BMVC22-workshop}.
\end{enumerate}
Object-Complete frames have the potential to become exceedingly large, so we employ a point subsampling strategy.

\begin{table}[htb]
    \centering
    \caption{Registration methods comparison for Object Complete Frame Generation on the ONCE Small split. We trained our best SECOND model in the semi-supervised setting with X-Ray Proficient Teacher on three different types of preprocessed data and compared results on ONCE validation set.}
    \begin{tabular}{l|l}
    \hline
    \textbf{Method} & \textbf{mAP} \\
    \hline
    Box Geometry & 67.88\\
    Greedy Grid  \cite{Bojanić-BMVC22-workshop} & 68.17 \\
    GeDi \cite{gedi} & \textbf{68.43}  \\
    \hline
    \end{tabular}

    \label{tab:PCR_comparison}
\end{table}

\begin{table}[!ht]
    \centering
 \caption{Comparison for supervised 3D Object Detection task on Waymo Open Dataset. We compare baseline models, their scaled versions and X-Ray distillation with default hyperparameters.  
 }
    \label{scaled_students}
   \resizebox{\columnwidth}{!}{
    \begin{tabular}{l|c|c|c}
        \toprule
        Model & mAP/mAPH L1 & mAP/mAPH L2 & \#params   \\
        \midrule

        SECOND X-Ray Teacher* & 85.1/70.3 & 75.1/64.7 & 5.3m \\
        \midrule
        SECOND & 67.2/63.1  & 61.0/57.2  & 5.3m  \\
        X-Ray SECOND & 67.0/62.8&  60.4/56.7  & 5.3m \\
        SECOND-Scaled & 66.8/62.7 & 59.4/56.1 & 6.2m  \\
        X-Ray SECOND-Scaled & \textbf{68.3/64.3} & \textbf{61.9/58.0} & 6.2m \\
        \midrule

        CenterPoint X-Ray Teacher* &  88.3/78.6 & 76.4/72.9 & 8.3m  \\
        \midrule
        CenterPoint & 74.4/71.7 & 68.2/65.8  & 8.3m  \\
        X-Ray CenterPoint  & 73.2/69.7 & 67.1/64.5  & 8.3m\\
        CenterPoint-Scaled & 74.1/71.5  & 67.9/65.3  & 9.2m \\
        X-Ray CenterPoint-Scaled & \textbf{75.2 /72.1} & \textbf{68.9/66.3} & 9.2m \\
        \midrule

        DSVT Pillar X-Ray Teacher* &89.3/79.7 & 79.1/73.4  &  8.6m \\
        \midrule
        DSVT Pillar & 79.5/77.1 & 73.2/71.0 & 8.6m  \\
        X-Ray DSVT Pillar & 79.2/76.7 & 72.6/70.3 & 8.6m \\
        DSVT Pillar-Scaled & 79.6/77.2 & 73.3/71.2 & 9.5m \\
        X-Ray DSVT Pillar-Scaled & \textbf{80.1/77.9} & \textbf{73.7/71.4} & 9.5m \\
        \bottomrule
    \end{tabular}}
\end{table}

 \subsection{Ablation Studies}

 \begin{figure}[]
    \centering
    \includegraphics[width=.85\columnwidth]{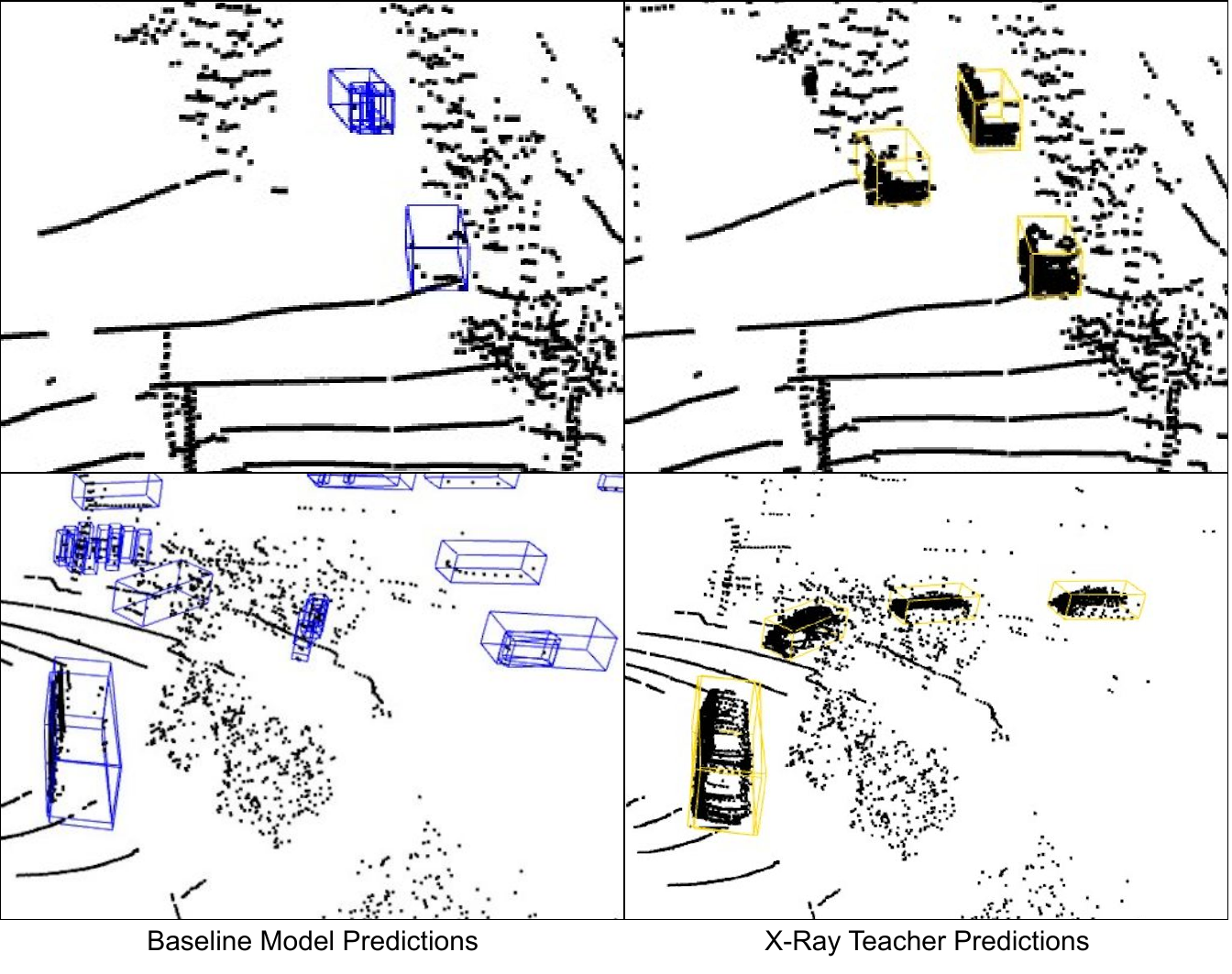}
    \caption{Visual comparison between noisy and poor baseline (original) 3D detector SECOND (left column) and our X-Ray Teacher that perceives Object-Complete frames. We compare two identical timestamps and view angles. The Baseline model fails to detect some objects while X-Ray Teacher does not. This explains why knowledge distillation is indeed beneficial and should improve models.}
    \label{x-ray-comp}
\end{figure}

\label{ablation_study}

In this section, we provide the comparison of different components of our approach and show how they affect the performance. Specifically, we prove the usefulness of teacher fine-tuning and compare Point Cloud Registration and distillation methods. We also analyse if dealing with sparsity and incomplete shapes really improves the performance.

First, we performed a detailed comparison of the semi-supervised performance of our X-Ray method, both with and without teacher fine-tuning using Object-Complete frames. We use Mean Teacher and Proficient Teacher with SECOND and CenterPointVoxel models trained on ONCE Small split. Table \ref{tab:finetuning_comparison} illustrates the substantial impact of teacher fine-tuning on performance, showing that neglecting this step results in a noticeable performance decline.

As we noted before, the reconstructed objects will not be perfect in the semi-supervised setting, which is precisely why we included Table \ref{tab:PCR_comparison}. PCR models trained on domain objects like cars, pedestrians, cyclists, etc., should improve our method even more.


As we mentioned earlier, we combine several techniques for the knowledge distillation: BEV features matching, heads output matching with simple regression and KL divergence and detection loss on the teachers predictions. We also compare partial solutions on the Waymo validation set with a SECOND model in Table \ref{tab:distillation_comparison}

\begin{table}[htb]
    \centering
    \caption{Comparison on NuScenes dataset. Our method improves baselines without scaling models due to the fact that NuScene's object-complete frames are less informative compared to Waymo.}
    \label{tab:nuscenes}    \begin{tabular}{l|l|l}
    \hline
    \textbf{Model}  & \textbf{mAP} & \textbf{NDS}\\
    \hline
    X-Ray Teacher* CBGS  & 77.1 & 72.4 \\
    \hline
    CBGS  & 50.0 & 59.2  \\
    X-Ray CBGS (ours) & \textbf{50.8} & \textbf{60.4} \\
    \hline\hline
    X-Ray Teacher* CenterPoint-Voxel   &79.5 & 77.1 \\
    \hline
    CenterPoint-Voxel  & 53.4 & 61.3  \\
    X-Ray CenterPoint-Voxel (ours) & \textbf{54.3} & \textbf{62.9} \\
    \hline\hline
    X-Ray Teacher* Transfusion-L   &81.3 & 78.2 \\
    \hline
    Transfusion-L  & 56.1 & 66.3  \\
    X-Ray Transfusion-L (ours) & \textbf{56.8} & \textbf{66.9} \\
    \hline
    \end{tabular}
\end{table}




\begin{table}[!htb]
 \centering
 \caption{Performance of X-Ray-powered Mean Teacher and Proficient Teacher methods in the semi-supervised setting using SECOND and CenterPoint baselines, ONCE validation set. Our approach consistently outperforms the state-of-the-art for Semi-Supervised 3D Object Detection in terms of mAP across all splits.}
 \label{results_combined}

 \begin{tabular}{l|l|l}
 \hline
 Method & SECOND & CenterPoint \\
                      
 \hline
 \multicolumn{3}{c}{\textbf{Train} (5k labeled samples)}        \\
 \hline
 Pretraining       &  63.22   & 64.41 \\
 \hline
 \multicolumn{3}{c}{\textbf{Small} (5k labeled + 100k unlabeled samples)}        \\
 \hline
 Mean Teacher       &  64.31  & 66.47 \\
 X-Ray MT (ours)    &  65.26 (\textcolor{green}{+0.95})  & 67.60 (\textcolor{green}{+1.13}) \\
 Proficient Teacher &  67.06  & 67.72 \\
 X-Ray PT (ours)    &  \textbf{68.43} (\textcolor{green}{+1.37})  & \textbf{68.76} (\textcolor{green}{+1.04}) \\
 \hline
 \multicolumn{3}{c}{\textbf{Medium} (5k labeled + 500k unlabeled samples)}      \\
 \hline
 Mean Teacher       &  64.73  & 66.87 \\
 X-Ray MT (ours)    &  65.62 (\textcolor{green}{+0.89})  & 67.78 (\textcolor{green}{+0.91}) \\
 Proficient Teacher &  67.49  & 68.54 \\
 X-Ray PT (ours)    &  \textbf{68.75}  (\textcolor{green}{+1.26})  & \textbf{69.96}  (\textcolor{green}{+1.42}) \\
 \hline
 \multicolumn{3}{c}{\textbf{Large} (5k labeled + 1M unlabeled samples)}      \\
 \hline
 Mean Teacher       &  65.03  & 67.45 \\
 X-Ray MT (ours)    &  65.97 (\textcolor{green}{+0.94})  & 68.17 (\textcolor{green}{+0.72}) \\
 Proficient Teacher &  67.89  & 69.68 \\
 X-Ray PT (ours)    &  \textbf{69.10} (\textcolor{green}{+1.21})  & \textbf{70.55} (\textcolor{green}{+0.87}) \\
 \hline
 \end{tabular}
 \end{table}

    
    



    
    

We evaluate various Point Cloud Registration (PCR) methods used in the Object Complete Frame Generation process. This analysis, detailed in Table \ref{tab:PCR_comparison}, is conducted on a ONCE Small split. The results indicate that superior PCR leads to the creation of less noisy objects, which in turn contributes to improved overall quality. However, it's important to note that methods, such as the Greedy Grid method \cite{Bojanić-BMVC22-workshop} and GeDi \cite{gedi}, are computationally more expensive. This introduces a trade-off between computational efficiency and the quality of the results, highlighting the need for a balanced approach in the selection of PCR methods.

\subsection{Model comparison}
\label{results}

\subsubsection{Supervised Learning}

We perform model comparisons using SECOND \cite{second}, CenterPoint-Voxel \cite{centerpoint}, and DSVT Pillar \cite{dsvt} on the Waymo dataset, and CBGS \cite{cbgs} and CenterPoint-Voxel on the NuScenes dataset. Additionally, we train scaled versions of these models without the X-Ray Teacher to show that the improvements in detection quality are due to the effectiveness of our method, not just an increase in the number of parameters. We scale Waymo students because of the extremely dense and complete point clouds, unlike NuScenes (see supplementary materials), where simpler data requires fewer parameters to learn meaningful feature representations. The results, presented in Tables \ref{scaled_students} and \ref{tab:nuscenes}, demonstrate that our approach consistently outperforms baseline models by 1-2 mAP.

\begin{table}[htb]
    \centering
    \caption{Student performance on ONCE validation small split with different tracking methods used in Objects Temporal Fusion}\label{tab:tracking_ablation}
\resizebox{.82\columnwidth}{!}{
    \begin{tabular}{l|l|l}
    \hline
    \textbf{Tracking method} & \textbf{SECOND} & \textbf{CenterPoint} \\
    \hline
    Greedy & 68.43 & 68.76\\
    \hline
    Kalman Filter + IoU & 68.57 & 68.91\\
    \hline
    ReID + Kalman Filter & \textbf{68.79} & \textbf{69.12}\\
    \hline
    \end{tabular}}
\end{table}

 \subsubsection{Semi-Supervised Learning}

To validate the effectiveness of our method in the context of pseudo-label-based semi-supervised learning, we perform a comparative analysis with the Mean Teacher \cite{MeanTeachers} and Proficient Teacher \cite{ProfTeachers} methods, which use the SECOND and CenterPointVoxel models. We compare the results obtained with and without the use of the X-Ray Teacher, as detailed in Table \ref{results_combined}. Our results show that the application of our approach consistently improves performance, yielding an improvement of 0.8-1.4 mAP. 

\section{Conclusion}


In our research, we proposed the innovative X-Ray Teacher framework, tailored to improve 3D Object Detection models in supervised and semi-supervised settings. Our extensive results have shown that this approach not only achieves state-of-the-art performance in the semi-supervised setting on the ONCE benchmark, but also consistently improves the quality of supervised models on NuScenes and Waymo Open Dataset. The main contributions of our work include the design of the X-Ray Teacher framework, the development of the Objects Temporal Fusion block for generating Object-Complete frames for data lacking ground truth tracking labels, and the demonstration of the potential of our method to improve the performance of any supervised model trained on sequential data. Future work will focus on optimizing the Objects Temporal Fusion block for more complex environments and exploring the integration of our framework with a broader range of model architectures and applications.

\section{Acknowledgements}
This research was supported in part through computational resources of HPC facilities at HSE University.

The work of I. Makarov on 3D object detection related work was supported by a grant for research centers in the field of artificial intelligence, provided by the Analytical Center in accordance with the subsidy agreement (agreement identifier 000000D730321P5Q0002) and the agreement with the Ivannikov Institute for System Programming of dated November 2, 2021 No. 70-2021-00142.
\clearpage

{\small
\bibliographystyle{ieee_fullname}
\bibliography{main}
}

\newpage

In these supplementary materials we provide detailed implementation insights: examples of both positive and negative Temporal Object Fusion, visualizations of NuScenes, Waymo and ONCE Object-Complete Frames, and additional models comparisons to provide a comprehensive overview of the work.

The supplementary materials are organized in the following way: 

\begin{enumerate}
    \item 
Section \ref{sec:default} demonstrates the performance improvements by the Semi-Supervised X-Ray Teacher method over the baseline model on the ONCE dataset, using the default training parameters provided by \cite{openpcdet2020}. This improvement is notable not only with the refined training parameters \cite{once_fix} utilized in main work but also when applying the default parameters that were commonly used in previous research \cite{3diou, ProfTeachers}. 
    \item Additional experiments and visualizations of the Object Temporal Fusion block can be found in Section \ref{sec:otf}. 
    \item Section \ref{sec:sup} shows detailed information on the Object-Complete frames preprocessing and training details of the Supervised X-Ray Teacher.

    \item Section \ref{sec:vis} presents a selection of randomly chosen Object-Complete and original frames from the NuScenes , Waymo, and ONCE datasets to provide a comparative view of how these frames typically differ.
\end{enumerate}

\section{Additional Semi-Supervised Evaluation}
\label{sec:default}

In the main text, we adopted the training parameters from our previous work \cite{once_fix} for training our model on the ONCE dataset. This particular paper revealed that the training parameters previously used to achieve state-of-the-art (SOTA) results were suboptimal during the pretraining stage, leading to an unfair comparison of semi-supervised methods. Nevertheless, we also present results using the default parameters proposed in \cite{once} to illustrate the robustness of our methods under various conditions and to enable a direct comparison of metrics with those reported in previous SOTA works. Other than this, we do not modify our method in any way.

\begin{table}[]
 \centering
 \caption{Comparison of the performance of X-Ray Teacher in semi-supervised setting vs other  methods using SECOND baseline model with a default configuration. Models were trained on different splits of unlabeled data (Small, Medium, Large) and evaluated on the ONCE validation split with  Mean Average Precision (mAP). The integration of X-Ray Teacher with the Mean Teacher and Proficient Teacher methods is referred to as X-Ray MT and X-Ray  PT, respectively. Higher metric values indicate superior model performance in 3D Object Detection. The best results are highlighted in \textbf{bold}. Values in parentheses indicate the performance difference between the original and X-Ray approaches. Our approach consistently outperforms the state-of-the-art for Semi-Supervised 3D Object Detection in terms of mAP across all splits.}
 \label{results_combined}

 \begin{tabular}{l|l}
 \hline
 Method & SECOND \\
                      
 \hline
 \multicolumn{2}{c}{\textbf{Train} (5k labeled samples)}        \\
 \hline
 Pretraining       &  51.89  \\
 \hline
 \multicolumn{2}{c}{\textbf{Small} (100k unlabeled samples)}        \\
 \hline
 Pseudo Label       &  51.22 (\textcolor{red}{-0.67})  \\
 Noisy Student       &  52.39 (\textcolor{green}{+0.50})  \\
 Mean Teacher       & 55.34  (\textcolor{green}{+3.45})\\
 SESS       &  53.39  (\textcolor{green}{+1.50})\\
 3DIoUMatch       &  53.81  (\textcolor{green}{+1.92})\\
 NoiseDet       &  58.00  (\textcolor{green}{+6.11})\\
 Proficient Teacher &  57.72  (\textcolor{green}{+5.83})\\
 X-Ray Teacher (ours)    &  \textbf{59.65}(\textcolor{green}{+7.76}) \\
 \hline
 \multicolumn{2}{c}{\textbf{Medium} (500k unlabeled samples)}      \\
 \hline
  Pseudo Label       &  50.40 (\textcolor{red}{-1.49})\\
 Noisy Student       &  55.34  (\textcolor{green}{+3.45})\\
 Mean Teacher       &  58.27  (\textcolor{green}{+6.38})\\
 SESS       &  55.79 (\textcolor{green}{+3.90}) \\
 3DIoUMatch       &  56.25  (\textcolor{green}{+4.36})\\
 NoiseDet       &  60.06  (\textcolor{green}{+8.17})\\
 Proficient Teacher &  59.89   (\textcolor{green}{+8.00})\\
 X-Ray Teacher (ours)    &  \textbf{62.42} (\textcolor{green}{+10.53})\\
 \hline
 \multicolumn{2}{c}{\textbf{Large} (1M unlabeled samples)}      \\
 \hline
  Pseudo Label      &  49.76  (\textcolor{red}{-2.13}) \\
 Noisy Student      &  56.37  (\textcolor{green}{+4.48})\\
 Mean Teacher       &  59.28 (\textcolor{green}{+7.39}) \\
 SESS               &  57.99 (\textcolor{green}{+6.10}) \\
 3DIoUMatch         &  57.07  (\textcolor{green}{+5.18})\\
 NoiseDet           &  61.16  (\textcolor{green}{+9.27})\\
 Proficient Teacher &  61.40   (\textcolor{green}{+9.51})\\
 X-Ray Teacher (ours) &  \textbf{63.57} (\textcolor{green}{+11.68})\\
 \hline
 \end{tabular}
 \end{table}


\begin{table}[]
    \centering
    \caption{Comparison of different registration methods for Object Complete Frame Generation. We used the default SECOND model in the semi-supervised setting with X-Ray Teacher (our modification of Mean Teacher without Exponential Moving Average) on three unlabeled splits ( Small, Medium, Large) processed with three different registration methods and evaluated it on the ONCE validation split with Mean Average Precision (mAP). Higher metric values indicate superior model performance in 3D Object Detection. The best results are highlighted in \textbf{bold}.  Our analysis shows that the choice of registration method has a noticeable impact on the performance of X-Ray Teacher. The GeDi registration method consistently outperforms the other techniques across all data splits, achieving the highest mAP scores. This underlines the importance of sophisticated registration techniques in the generation of more accurate and complete point clouds.}
    \begin{tabular}{l|l}
    \hline
    \multicolumn{2}{c}{\textbf{Small} }      \\
    \hline
    \textbf{Method} & \textbf{mAP} \\
    \hline
    Box Geometry & 59.04\\
    Greedy Grid  \cite{Bojanić-BMVC22-workshop} & 59.11 \\
    GeDi (our preprocessing)\cite{gedi} & \textbf{59.65}  \\
    \hline
    \multicolumn{2}{c}{\textbf{Medium} }      \\
    \hline
    Box Geometry & 62.03\\
    Greedy Grid  \cite{Bojanić-BMVC22-workshop} & 62.31 \\
    GeDi (our preprocessing) \cite{gedi} & \textbf{62.42}  \\
    \hline
    \multicolumn{2}{c}{\textbf{Large} }      \\
    \hline
    Box Geometry & 62.56\\
    Greedy Grid  \cite{Bojanić-BMVC22-workshop} & 62.81 \\
    GeDi (our preprocessing) \cite{gedi} & \textbf{63.57}  \\
    \end{tabular}

    \label{pcrs}
\end{table}

\section{Object Temporal Fusion}
\label{sec:otf}

\begin{figure*}[]
    \centering
    \includegraphics[width=\linewidth]{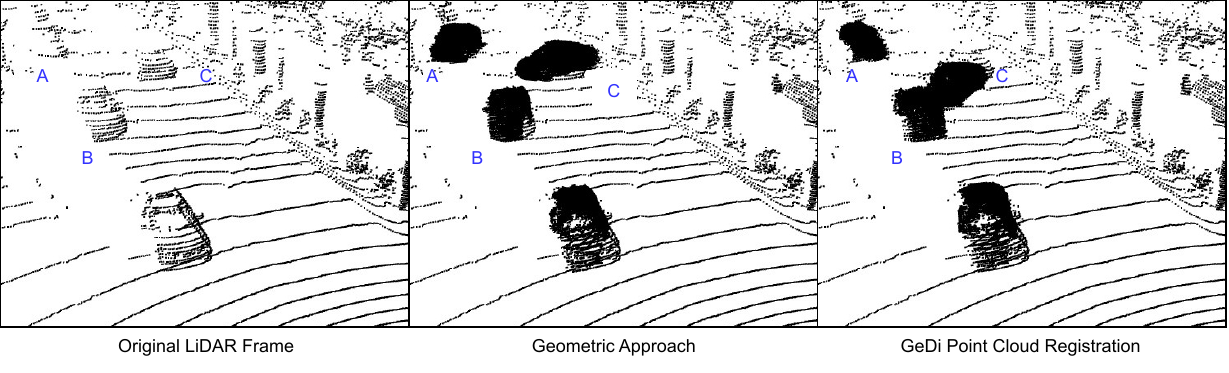}
    \caption{Visual comparison between the geometric approach (box alignment) and the GeDi \cite{gedi} Point Cloud Registration technique within the Objects Temporal Fusion module. We evaluate their effectiveness in merging different views of an object into a unified point cloud. Although both methods encounter difficulties with certain objects such as car C, the GeDi method generally provides superior alignment for other vehicles, like cars A and B. This indicates that while the GeDi technique is more effective at aligning object points, it also highlights the necessity for domain adaptation to prevent the errors, as observed with car C. The illustration underscores the potential of advanced registration methods in enhancing object detection while also pointing to the need for further refinement to ensure consistent accuracy across all objects.}
    \label{pcr_vis}
\end{figure*}

In this section, we provide a bit more detailed analysis and comparison of various Point Cloud Registration (PCR) methods utilized in the Object Temporal Fusion block. Initially, we replicated an experiment to compare different registration methods using default training parameters: the naive geometric approach, Greedy Grid \cite{Bojanić-BMVC22-workshop}, and GeDi \cite{gedi}, as outlined in Table \ref{pcrs}. The last method's superior performance is largely due to its ability to minimize noise effects in the rotation and coordinates of boxes. For instance, the SECOND model often produces noisy boxes on unlabeled samples, leading to misaligned objects with basic alignment methods. The comparative visualizations are shown in Figure \ref{pcr_vis}. From these visualizations, it is evident that both the geometric and GeDi methods encounter challenges in certain scenarios, such as with car C, while cars A and B exhibit notably better alignment using the GeDi method. This enhanced alignment is a result of advanced PCR process, which effectively reduces noise from the 3D detector. However, since registration applies corrections sequentially, an error in one iteration could lead to amplified errors in subsequent ones. This is why GeDi fails with car C but succeeds in transforming cars A and B into well-aligned, nearly complete objects, maintaining their original rotation. The Greedy Grid method is omitted in this analysis as its performance improvement is not on par with that of GeDi, and it generally aligns with the geometric approach in terms of overall quality.

\section{Supervised Setting Implementation}
\label{sec:sup}
This section discusses the specifics of Object-Completion Sampling, a technique used to optimize the processing of shape-complete objects, and provides insight into the training methodology for the Supervised X-Ray Teacher.
\subsection{Object-Completion}
Object-Completion is designed to utilize all available frames within a scene to create the most comprehensive representation of an object from every possible viewing angle. However, in scenarios where scenes are extended and contain numerous objects that remain within view over many frames, the size of the Object-Complete Frames can become exceedingly large. Specifically, in the Waymo dataset \cite{waymo}, a single frame might exceed 150MB, which can significantly slow down the training process. Moreover, having an excessive number of points for one scene can be unnecessary. To address this, we employ a straightforward sampling strategy for the Waymo dataset: we split the object-complete cloud into two distinct parts — the original frame and all newly added points. We then sample a volume 1.5 times the size of the original cloud from the new points and concatenate these samples with the original points. As for the NuScenes dataset, which typically generates much smaller Object-Complete frames, we do not implement any sampling strategy.

\subsection{Training}

\textbf{Teacher.} We train teacher networks from scratch: it never saw any original point cloud, only Object-Complete ones. We set default hyperparameters except for batch size - we match total number of samples processed simultaneously, so we make it equal 8.
\textbf{Knowledge Distillation.} 
We train students with the same configuration as if we were training teacher or original model. There's definitely a better configuration for our case and further researchers might also improve our results by just adjusting some hyperparameters.

\section{Visualization}
\label{sec:vis}

\begin{figure}[]

    \centering
    \includegraphics[width=\columnwidth]{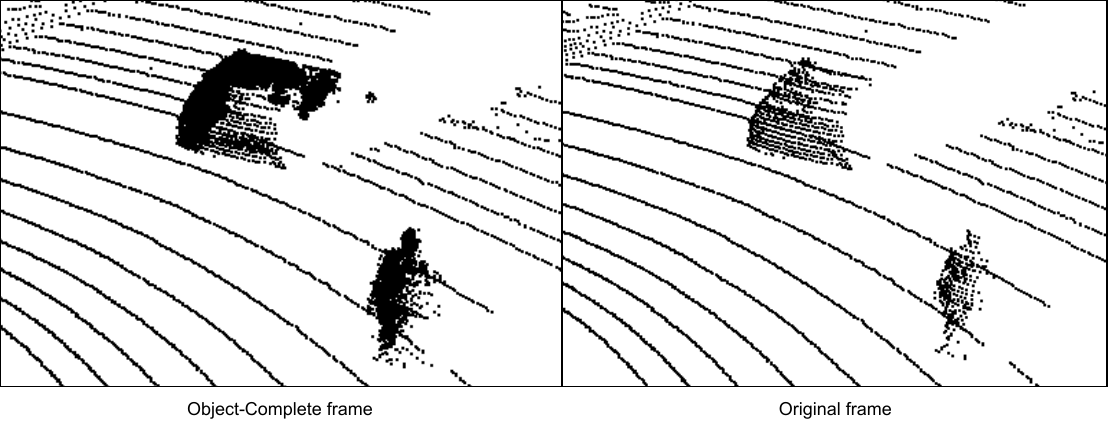}
    
    \caption{This figure provides a comparison of randomly selected objects from the ONCE validation set, showing the difference between Object-Complete frames (left) and original frames (right). The Object-Complete frame demonstrate the enhanced detail achieved through our frame generation process, which collects comprehensive point cloud data to construct a more complete representation of each object. This enhanced representation helps reduce the ambiguity typically associated with sparse LiDAR data, resulting in more accurate object detection.}
    \label{once_comp}
\end{figure}

\begin{figure*}[]

    \centering
    \fbox{\includegraphics[width=\columnwidth]{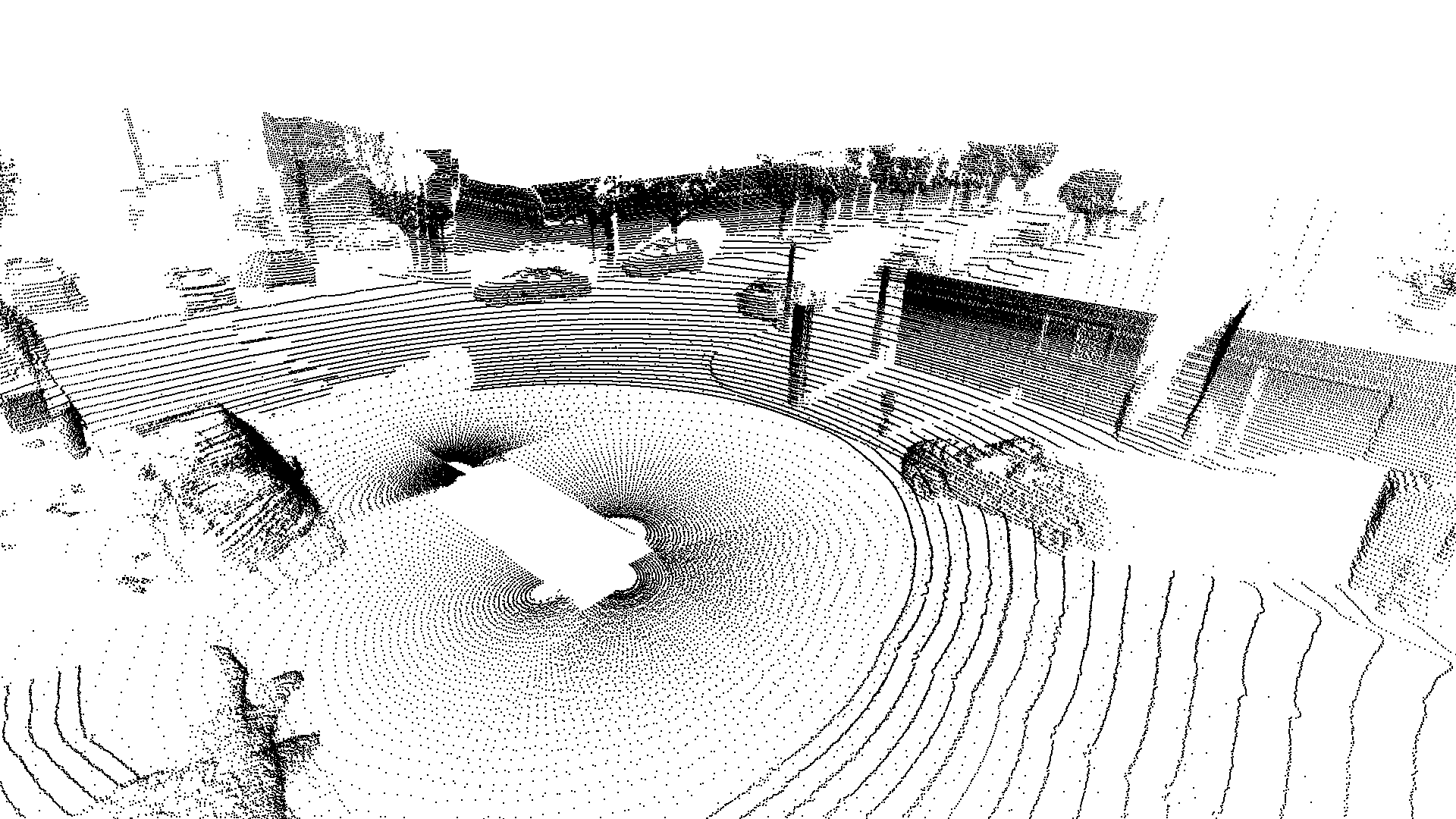}}
    \fbox{\includegraphics[width=\columnwidth]{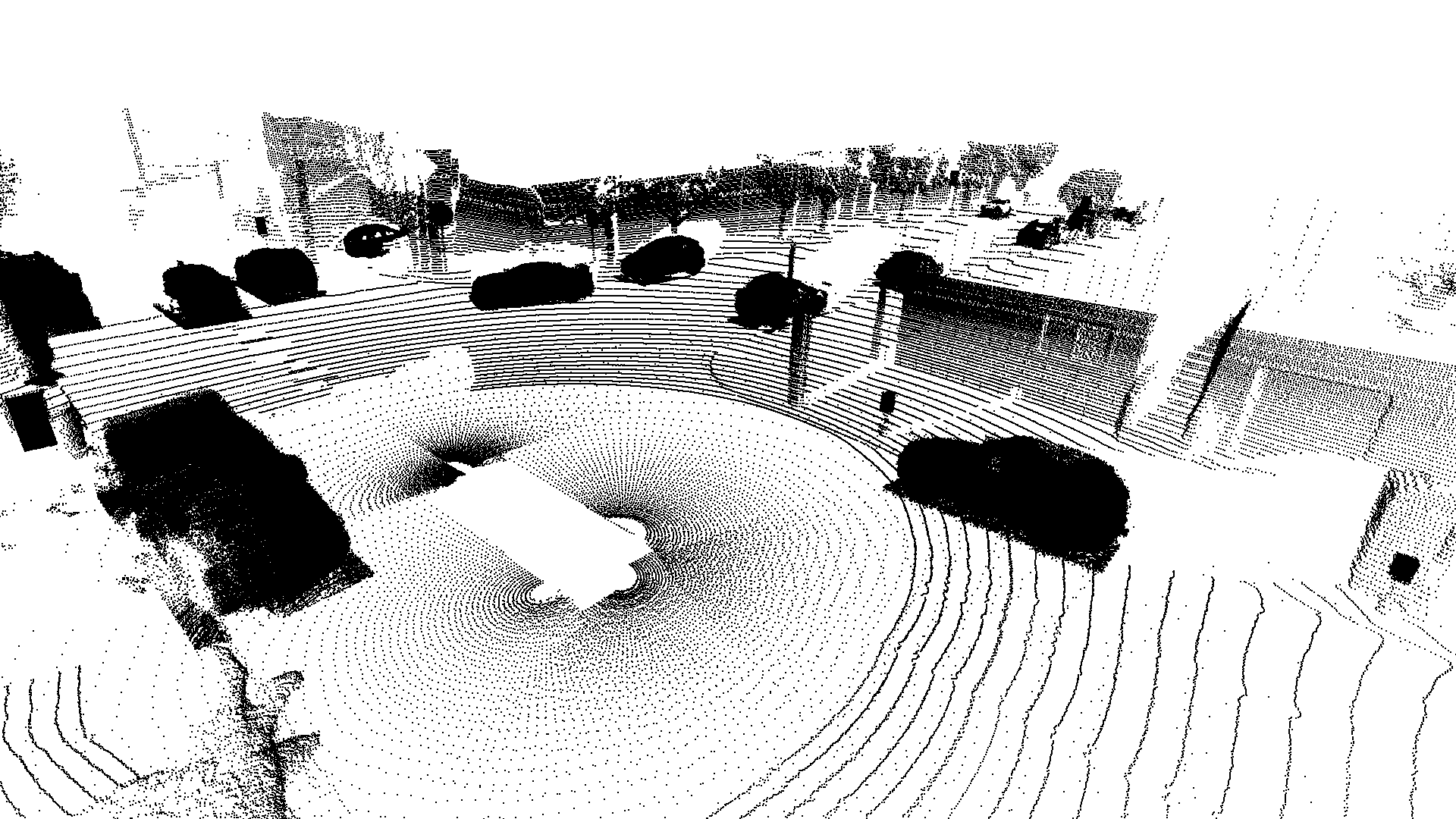}}
    \fbox{\includegraphics[width=\columnwidth]{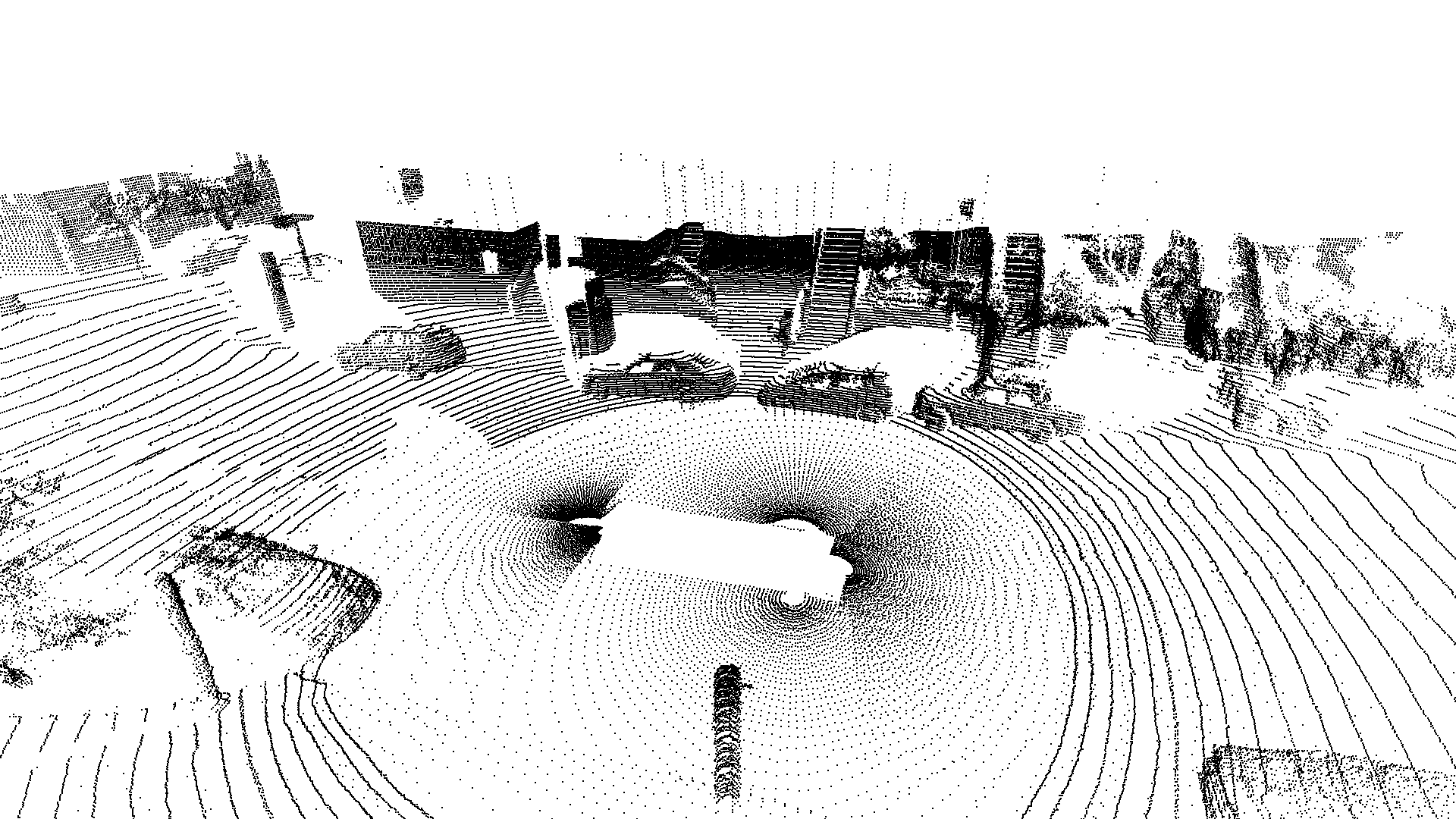}}
    \fbox{\includegraphics[width=\columnwidth]{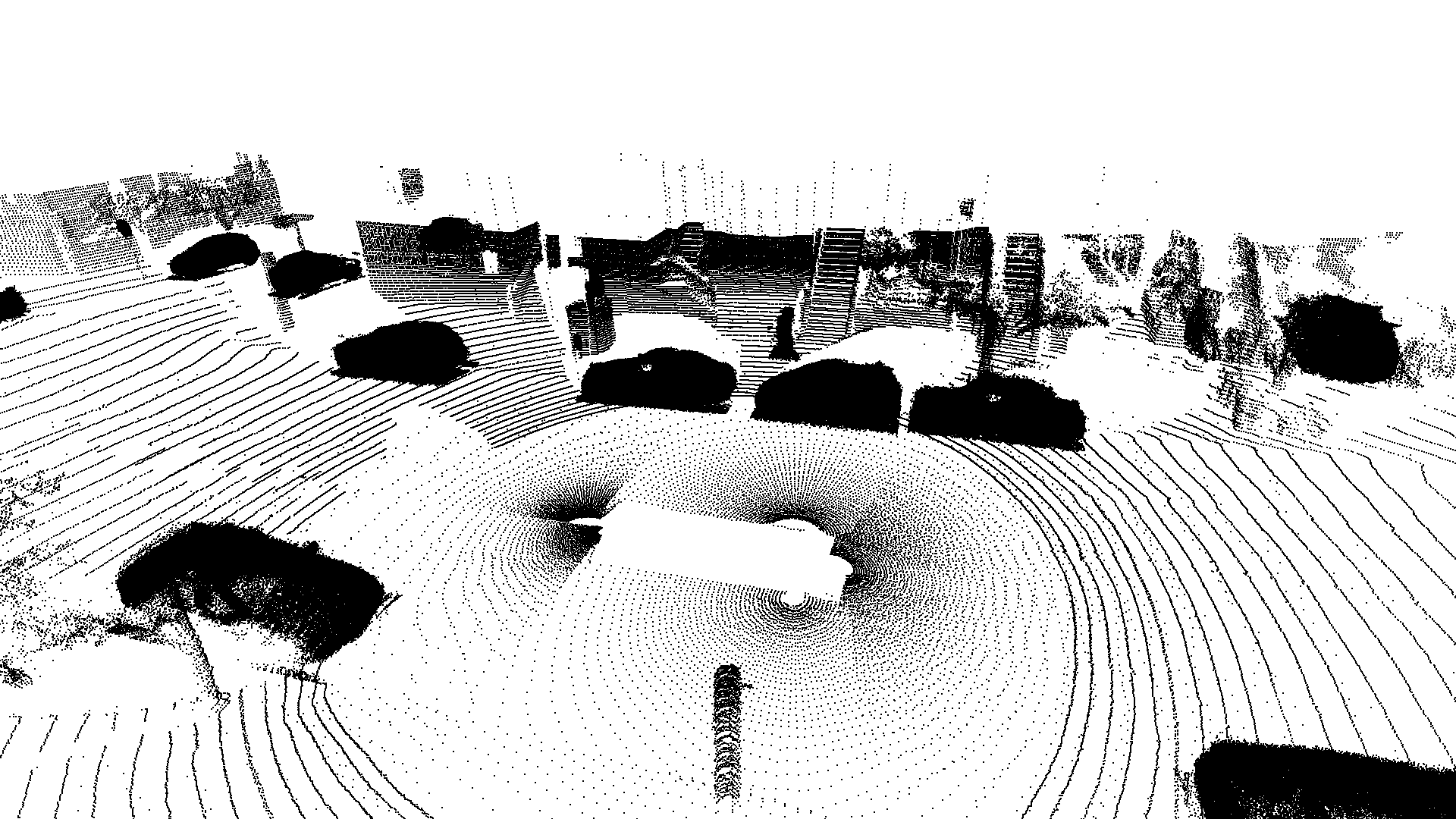}}
    \fbox{\includegraphics[width=\columnwidth]{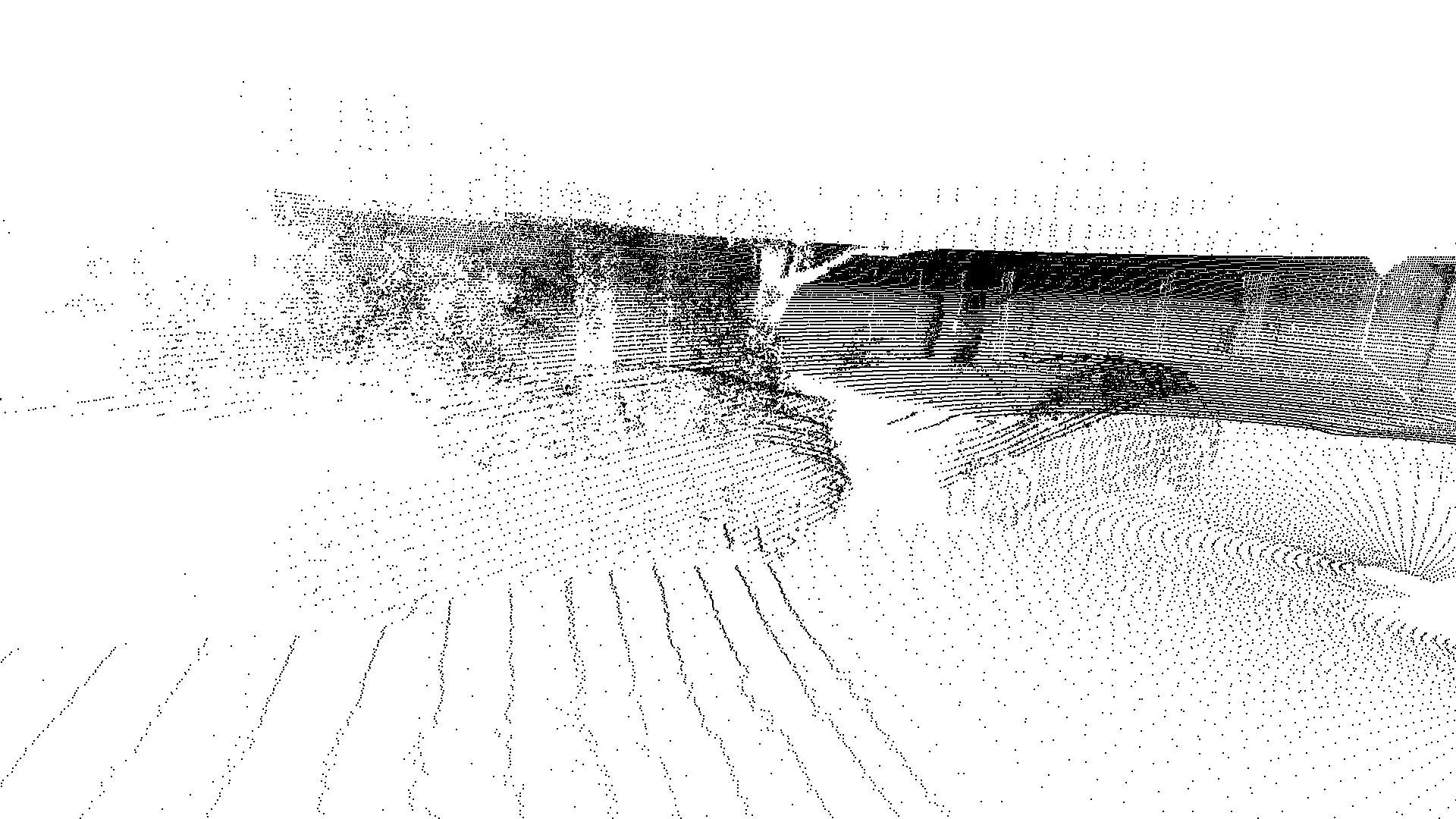}}
    \fbox{\includegraphics[width=\columnwidth]{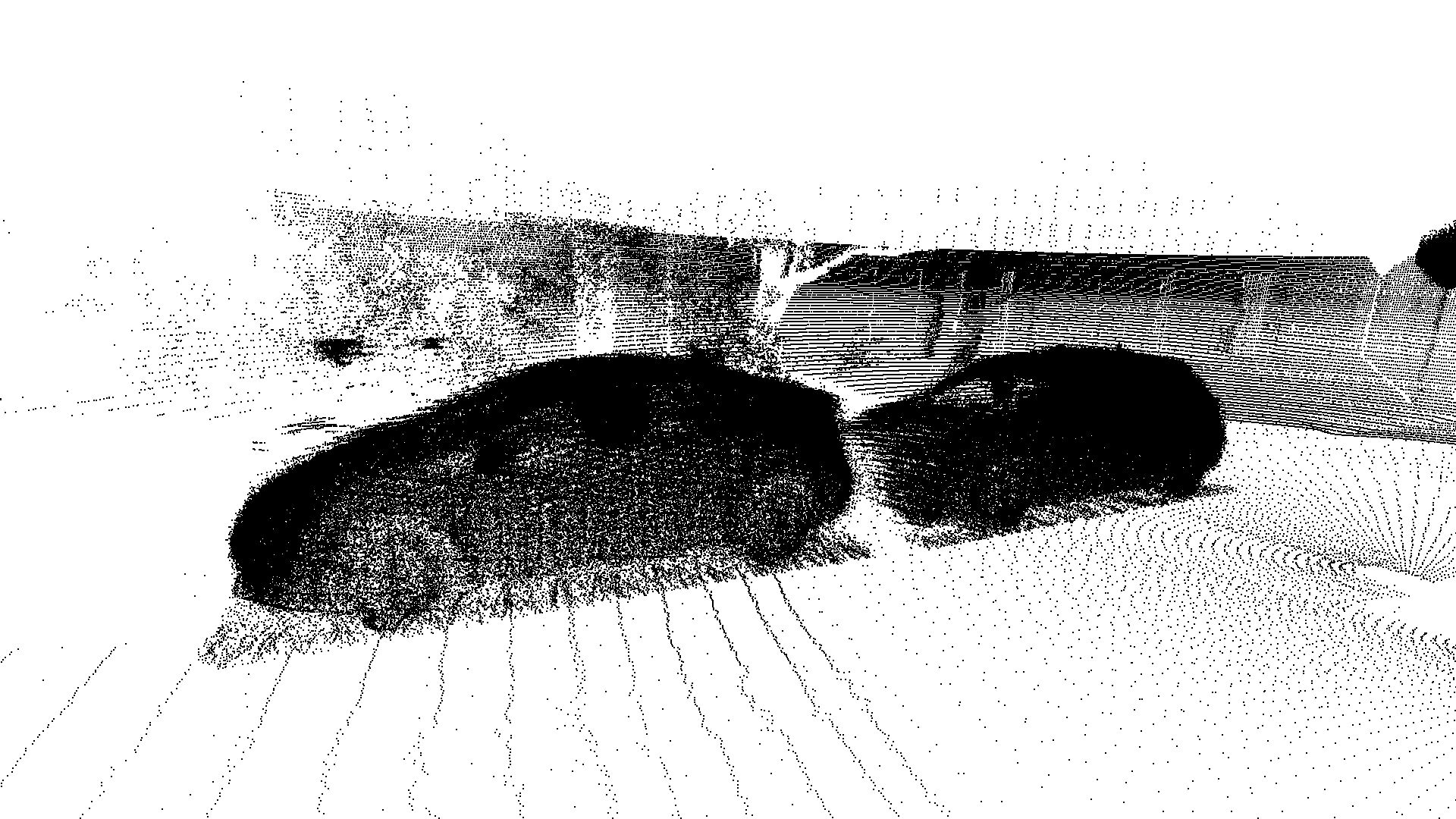}}
    \caption{Visual comparison between original (left) and Object-Complete (right) frames from the Waymo dataset. This figure shows how Object-Complete Frame Generation enriches point cloud data. This enhancement significantly diminishes sparsity and occlusions, thereby reducing ambiguity and making shape-complete objects easier to detect.}
    \label{x-ray-comp-waymo}
\end{figure*}

\begin{figure*}[]

    \centering
    \fbox{\includegraphics[width=\columnwidth]{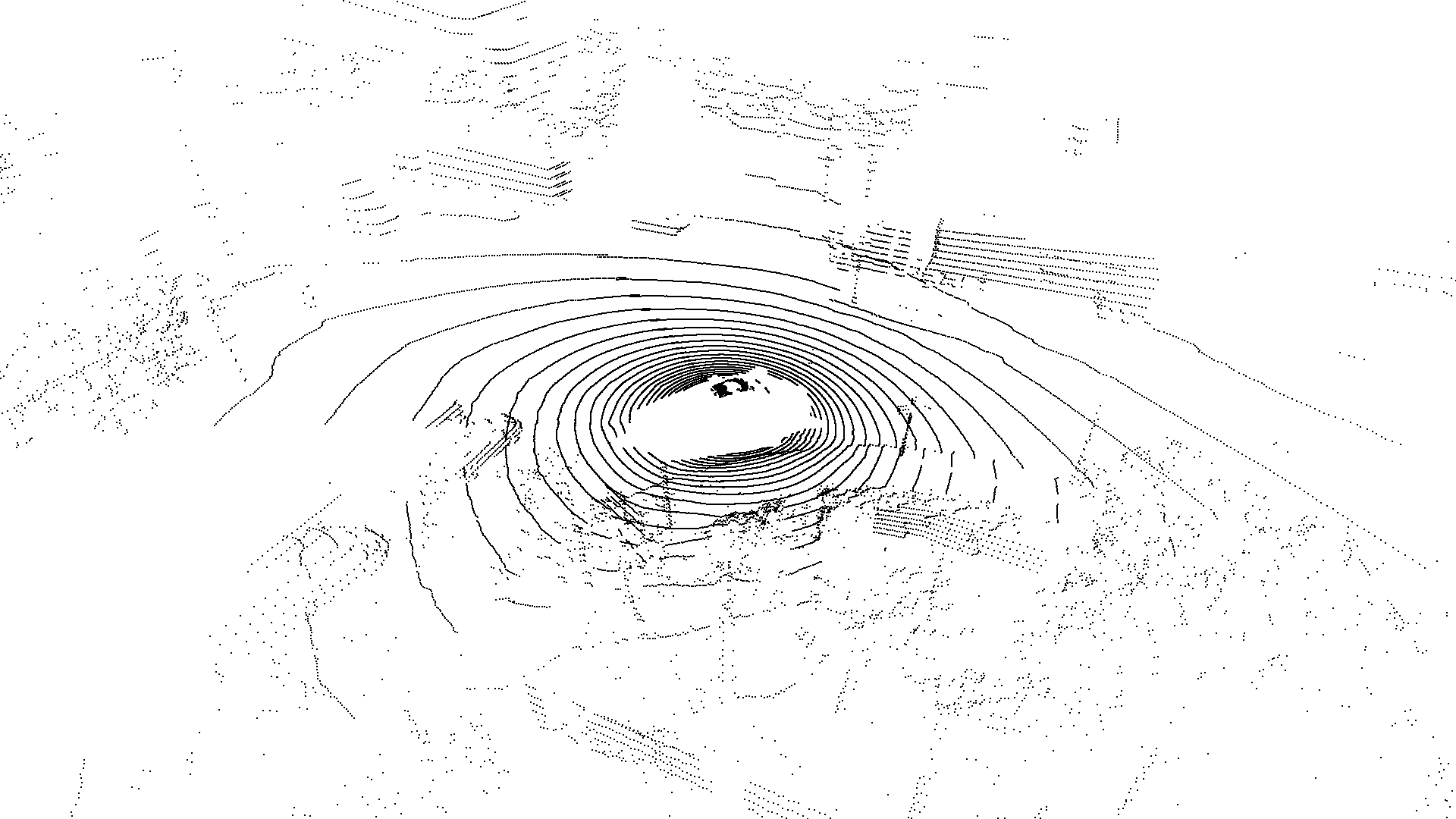}}
    \fbox{\includegraphics[width=\columnwidth]{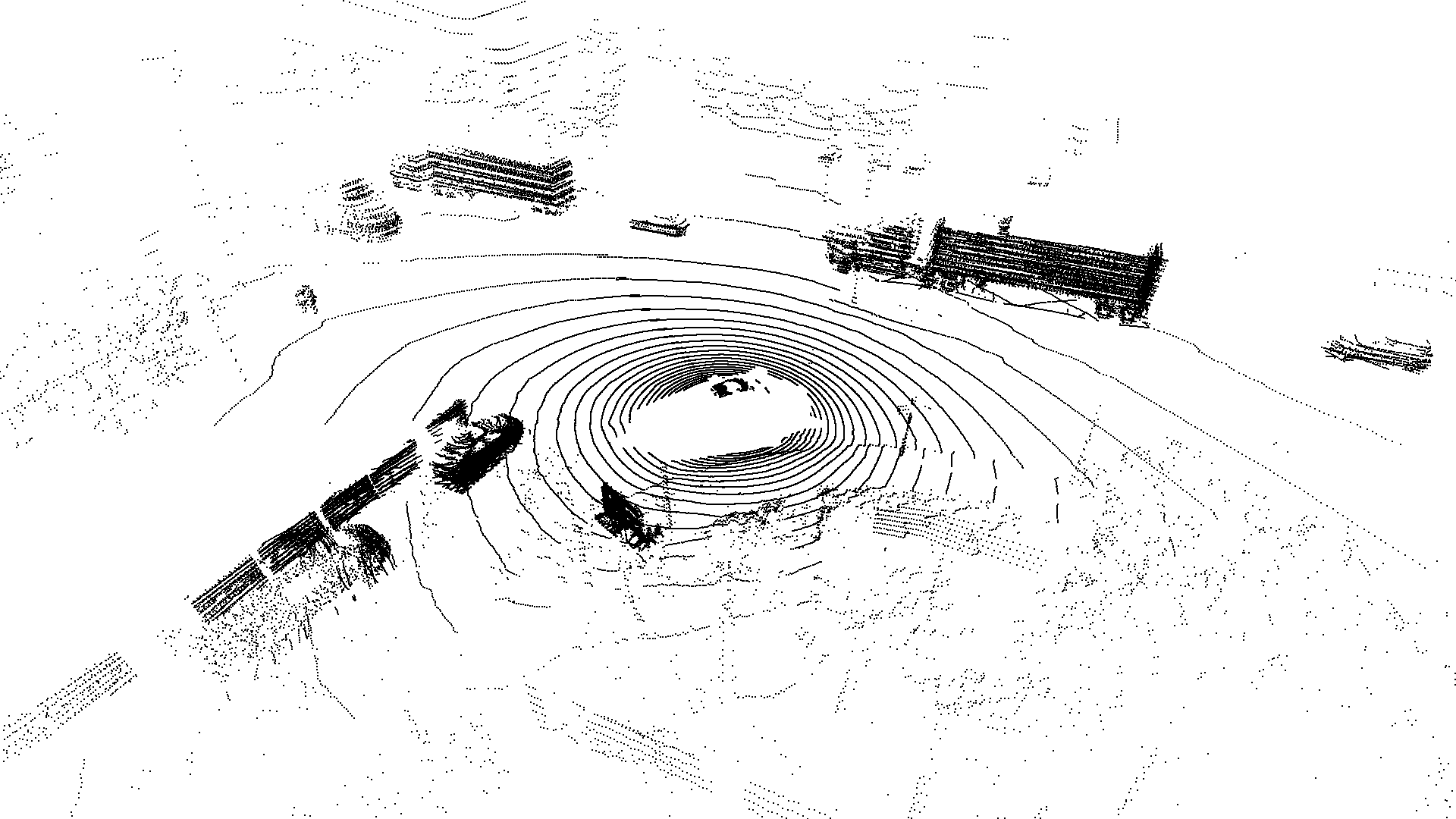}}
    \fbox{\includegraphics[width=\columnwidth]{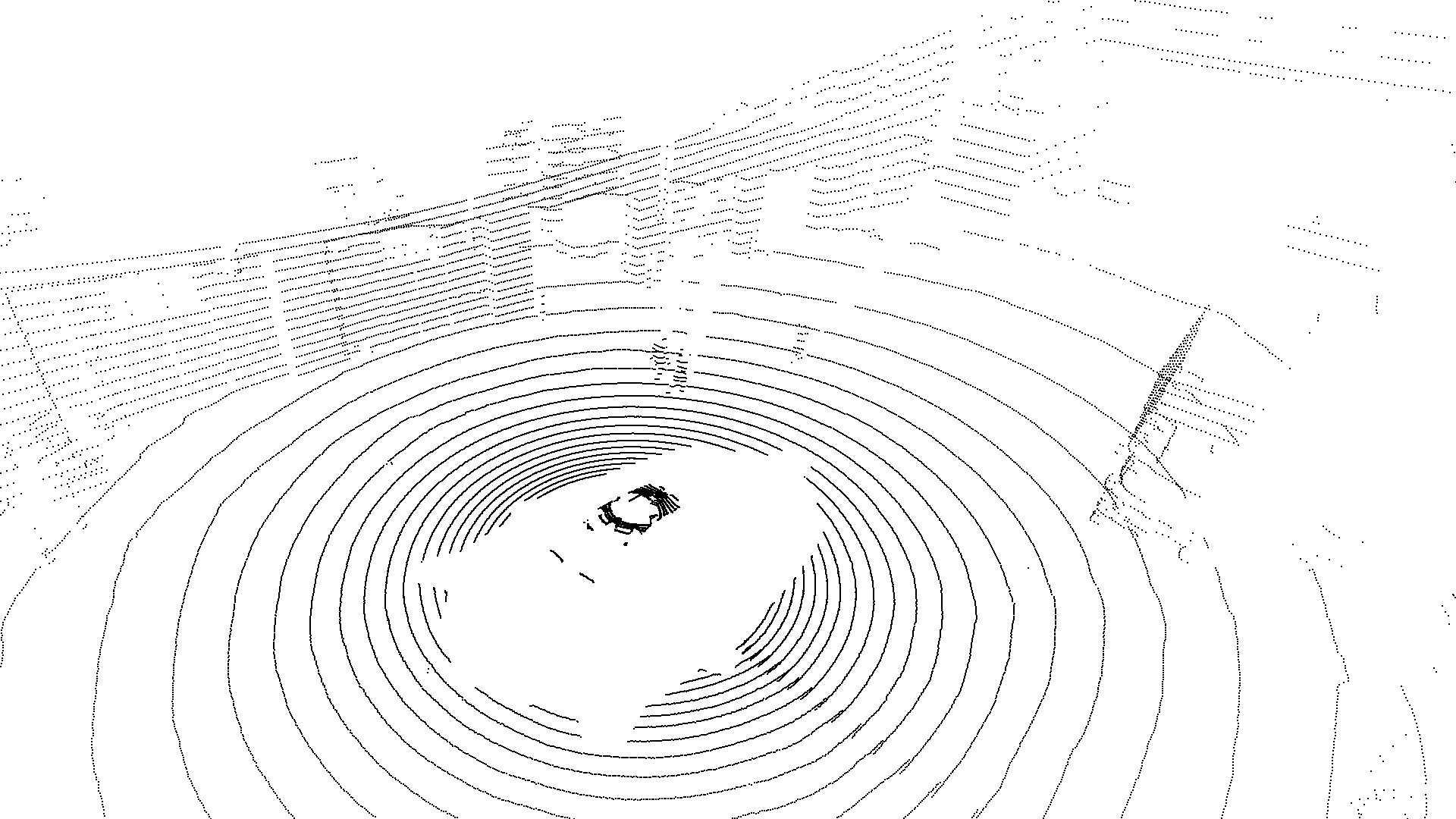}}
    \fbox{\includegraphics[width=\columnwidth]{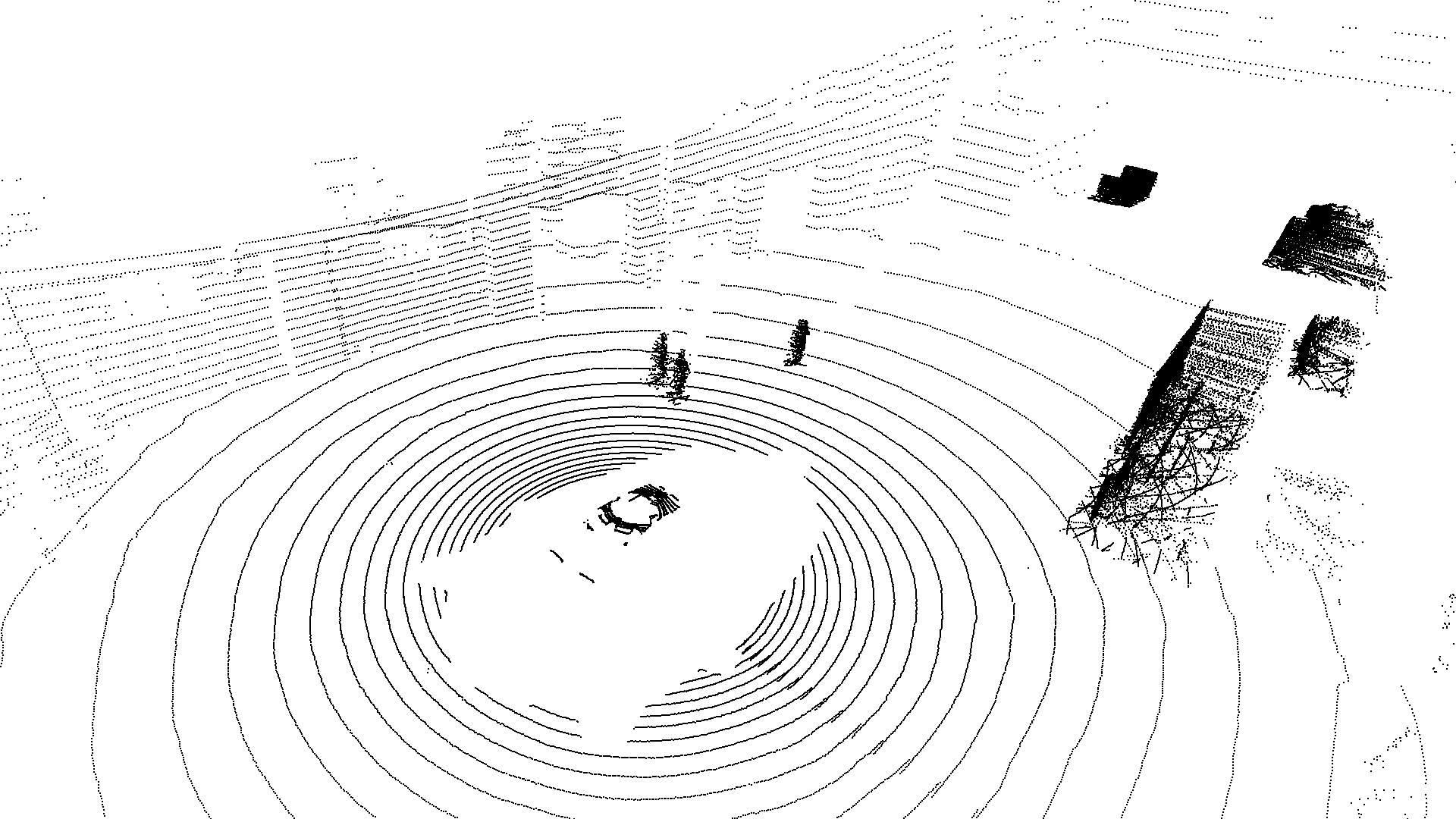}}
    \fbox{\includegraphics[width=\columnwidth]{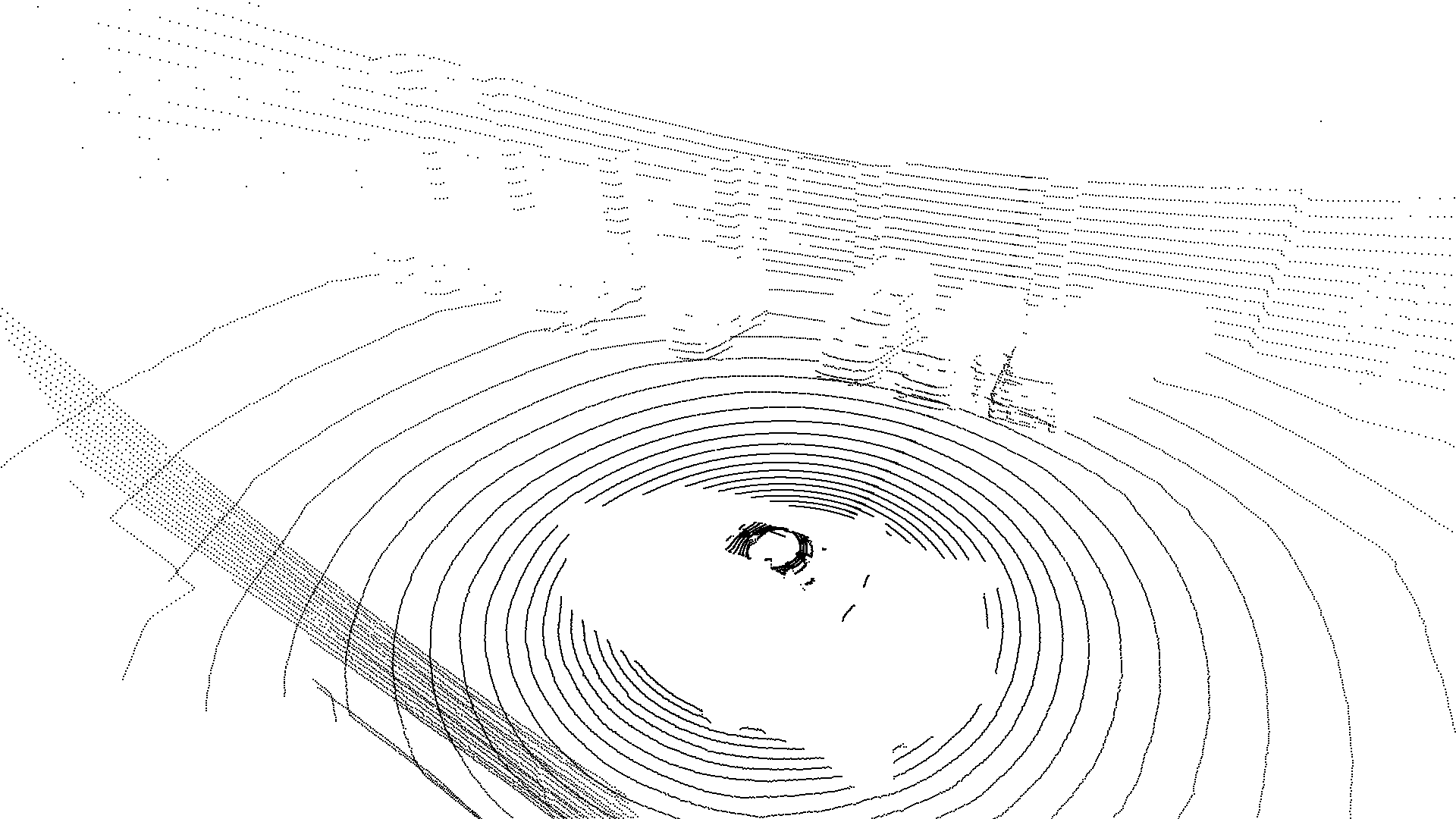}}
    \fbox{\includegraphics[width=\columnwidth]{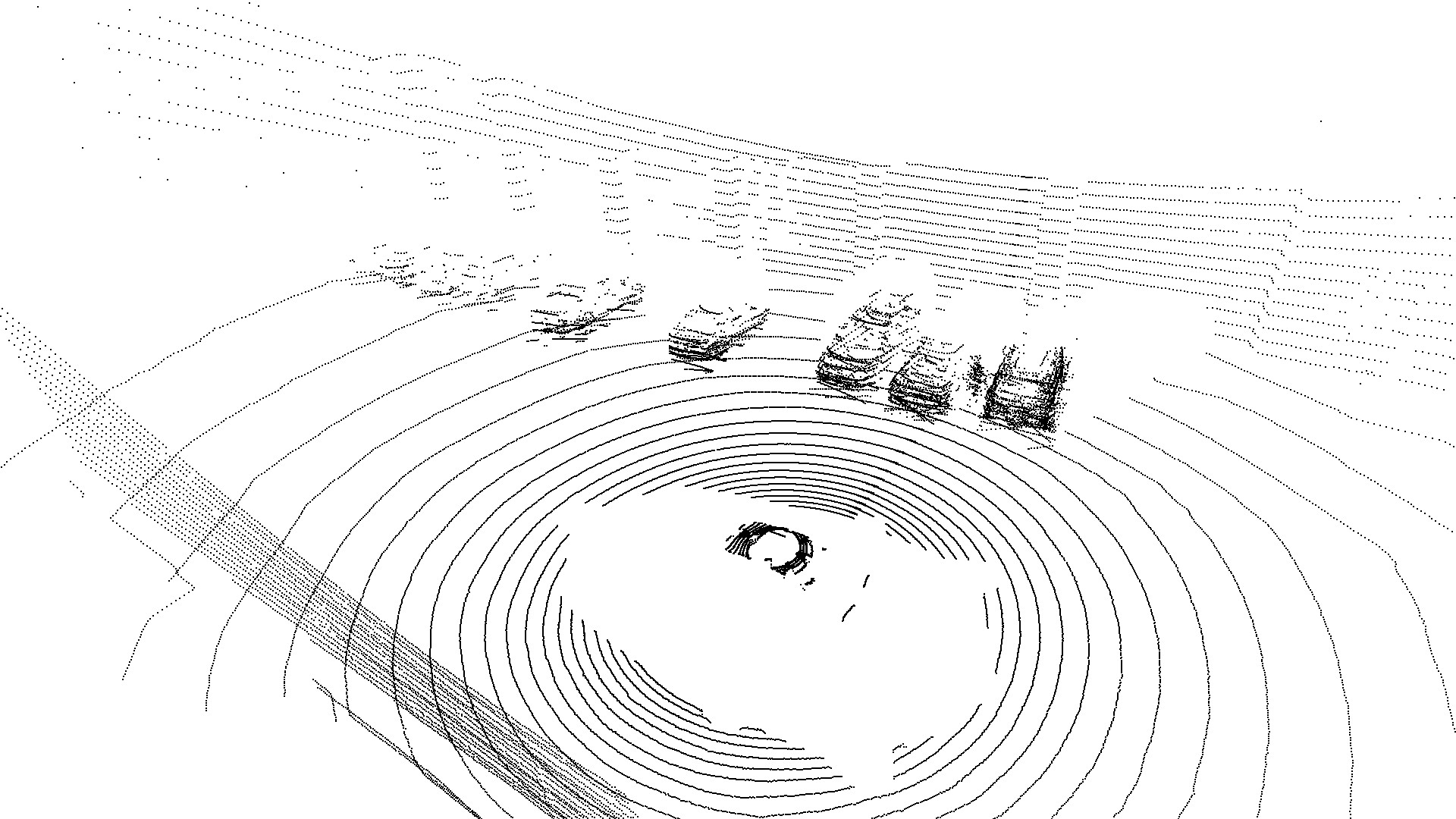}}
    \fbox{\includegraphics[width=\columnwidth]{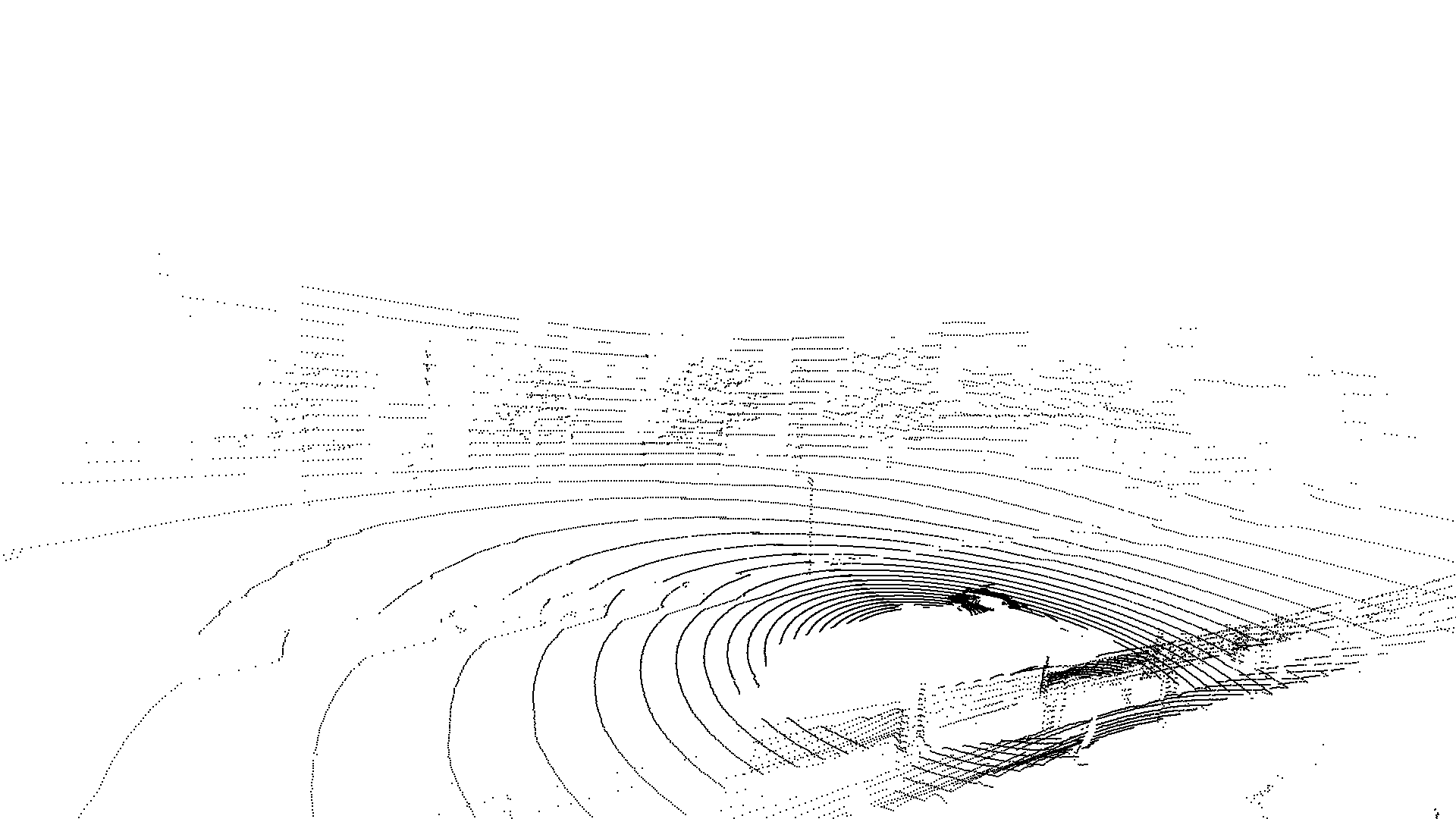}}
    \fbox{\includegraphics[width=\columnwidth]{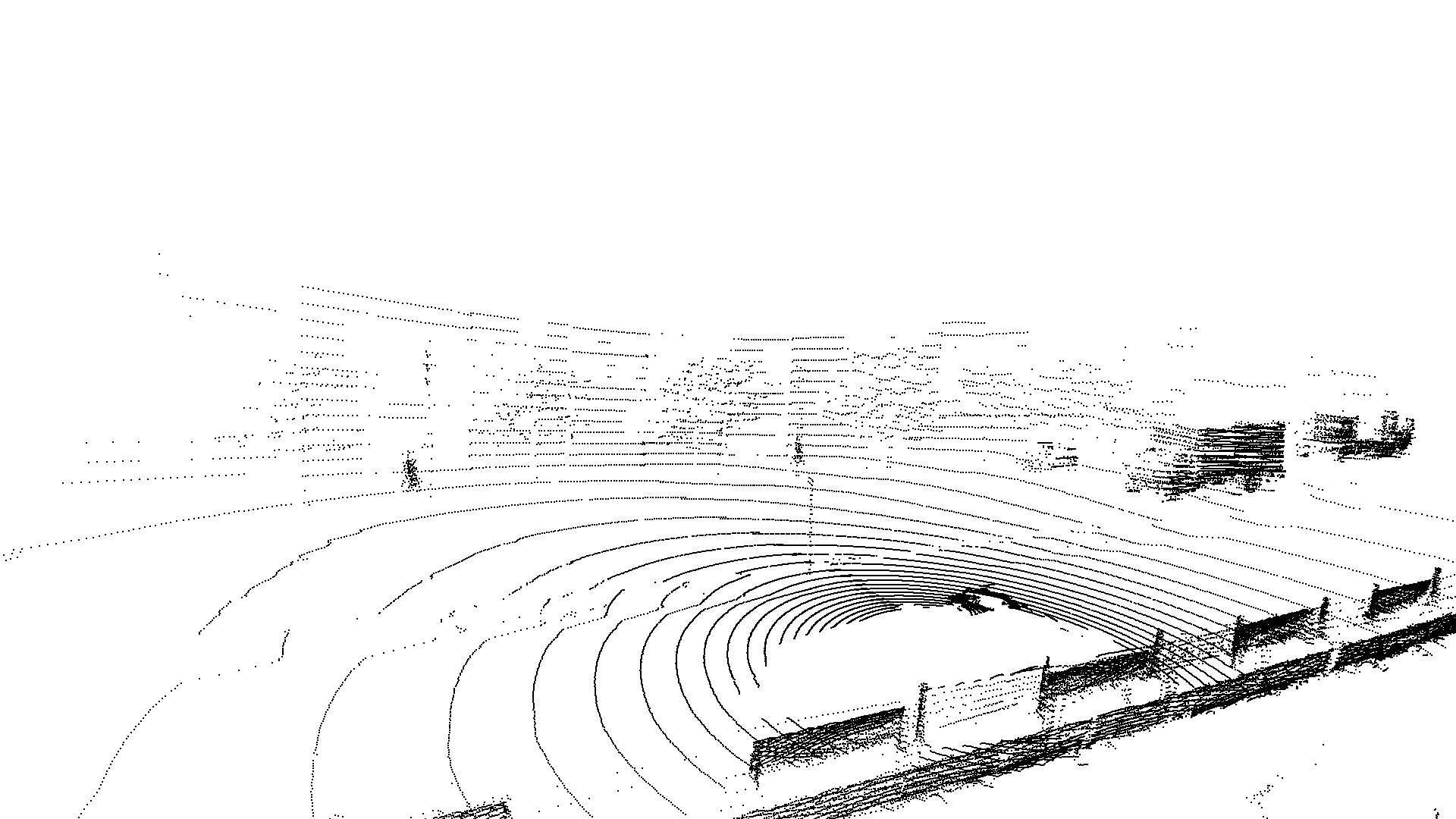}}
    \caption{Visual comparison between original (left) and Object-Complete (right) frames from the NuScenes dataset. This figure shows how Object-Complete Frame Generation enriches point cloud data. This enhancement significantly diminishes sparsity and occlusions, thereby reducing ambiguity and making shape-complete objects easier to detect.}
    \label{x-ray-comp-nuscenes}
\end{figure*}

In this section, we present a series of visual comparisons that highlight the effectiveness of our Object-Complete Frame Generation process. Figures \ref{once_comp}, \ref{x-ray-comp-waymo}, and \ref{x-ray-comp-nuscenes} illustrate the pronounced differences between original frames and those enhanced by Object-Complete Frame Generation, using examples from various real-world autonomous driving datasets. These comparisons visually demonstrate the advantages of our method in terms of data richness and object detection clarity. By significantly reducing sparsity and occlusions, the generated Object-Complete frames offer a more accurate and unambiguous representation of the scene, as can be seen in the enhanced details of the objects. These visualizations serve to illustrate the practical benefits of our approach, reinforcing the validity of our contributions to the field of LiDAR-based 3D object detection.

\end{document}